\documentclass{article}

\usepackage{microtype}
\usepackage{booktabs} 

\usepackage[accepted]{styles/icml2019}

\usepackage{url}
\usepackage{graphicx}
\usepackage{grffile}
\usepackage{hyperref}
\usepackage{subcaption}
\usepackage{amsmath, amssymb, bm, bbm}
\usepackage{physics}
\usepackage{mathrsfs}
\usepackage{mathtools}
\usepackage[scr=boondoxo]{mathalfa}
\usepackage{cleveref}
\usepackage{wrapfig}
\usepackage{color}
\usepackage{lipsum}
\usepackage{multirow}
\usepackage[table,svgnames]{xcolor}
\usepackage{here}

\newcommand{\cellbest}{\bf}

\DeclareMathOperator{\arccosh}{arccosh}
\DeclareMathOperator{\PT}{PT}
\DeclareMathOperator{\proj}{proj}

\icmltitlerunning{A Wrapped Normal Distribution on Hyperbolic Space for Gradient-Based Learning}

\begin{document}

\twocolumn[
\icmltitle{A Wrapped Normal Distribution on Hyperbolic Space for Gradient-Based Learning}




\begin{icmlauthorlist}
\icmlauthor{Yoshihiro Nagano}{utokyo}
\icmlauthor{Shoichiro Yamaguchi}{pfn}
\icmlauthor{Yasuhiro Fujita}{pfn}
\icmlauthor{Masanori Koyama}{pfn}
\end{icmlauthorlist}

\icmlaffiliation{utokyo}{Department of Complexity Science and Engineering, The University of Tokyo, Japan}
\icmlaffiliation{pfn}{Preferred Networks, Inc., Japan}

\icmlcorrespondingauthor{Yoshihiro Nagano}{nagano@mns.k.u-tokyo.ac.jp}

\icmlkeywords{Machine Learning, ICML}

\vskip 0.3in
]



\printAffiliationsAndNotice{}  

\begin{abstract}
Hyperbolic space is a geometry that is known to be well-suited for representation learning of data with an underlying hierarchical structure.
In this paper, we present a novel hyperbolic distribution called \textit{pseudo-hyperbolic Gaussian}, a Gaussian-like distribution on hyperbolic space whose density can be evaluated analytically and differentiated with respect to the parameters.
Our distribution enables the gradient-based learning of the probabilistic models on hyperbolic space that could never have been considered before.
Also, we can sample from this hyperbolic probability distribution without resorting to auxiliary means like rejection sampling.
As applications of our distribution, we develop a hyperbolic-analog of variational autoencoder and a method of probabilistic word embedding on hyperbolic space.
We demonstrate the efficacy of our distribution on various datasets including MNIST, Atari 2600 Breakout, and WordNet.
\end{abstract}

\section{Introduction}
\label{sec:intro}

Recently, hyperbolic geometry is drawing attention as a powerful geometry to assist deep networks in capturing fundamental structural properties of data such as a hierarchy.
Hyperbolic attention network \cite{Gulcehre2018} improved the generalization performance of neural networks on various tasks including machine translation by
imposing the hyperbolic geometry on several parts of neural networks.
Poincar\'e embeddings \cite{Nickel2017} succeeded in learning a parsimonious representation of symbolic data by embedding the dataset into Poincar\'e balls.

\begin{figure}[!h]
  \centering
  \begin{tabular}{c}
    \begin{minipage}[t]{0.55\hsize}
      \centering
      \subcaption{A tree representation of the training dataset}
      \vspace{-0.1cm}
      \includegraphics[width=\linewidth]{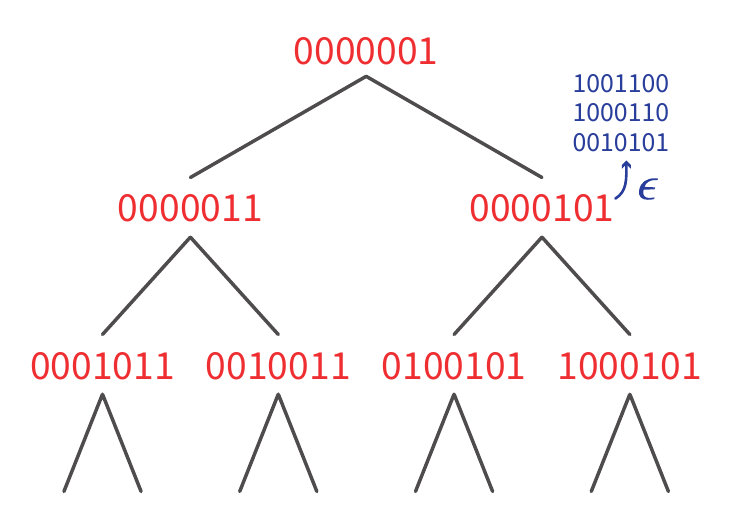}
      \label{fig:binary_tree_schema}
    \end{minipage}
    \end{tabular}
    \\
    \vspace{-0.25cm}
    \begin{tabular}{cc}
    \begin{minipage}[t]{0.47\hsize}
      \centering
      \subcaption{Vanilla VAE ($\beta=1.0$)}
      \vspace{-0.25cm}
      \includegraphics[width=\linewidth]{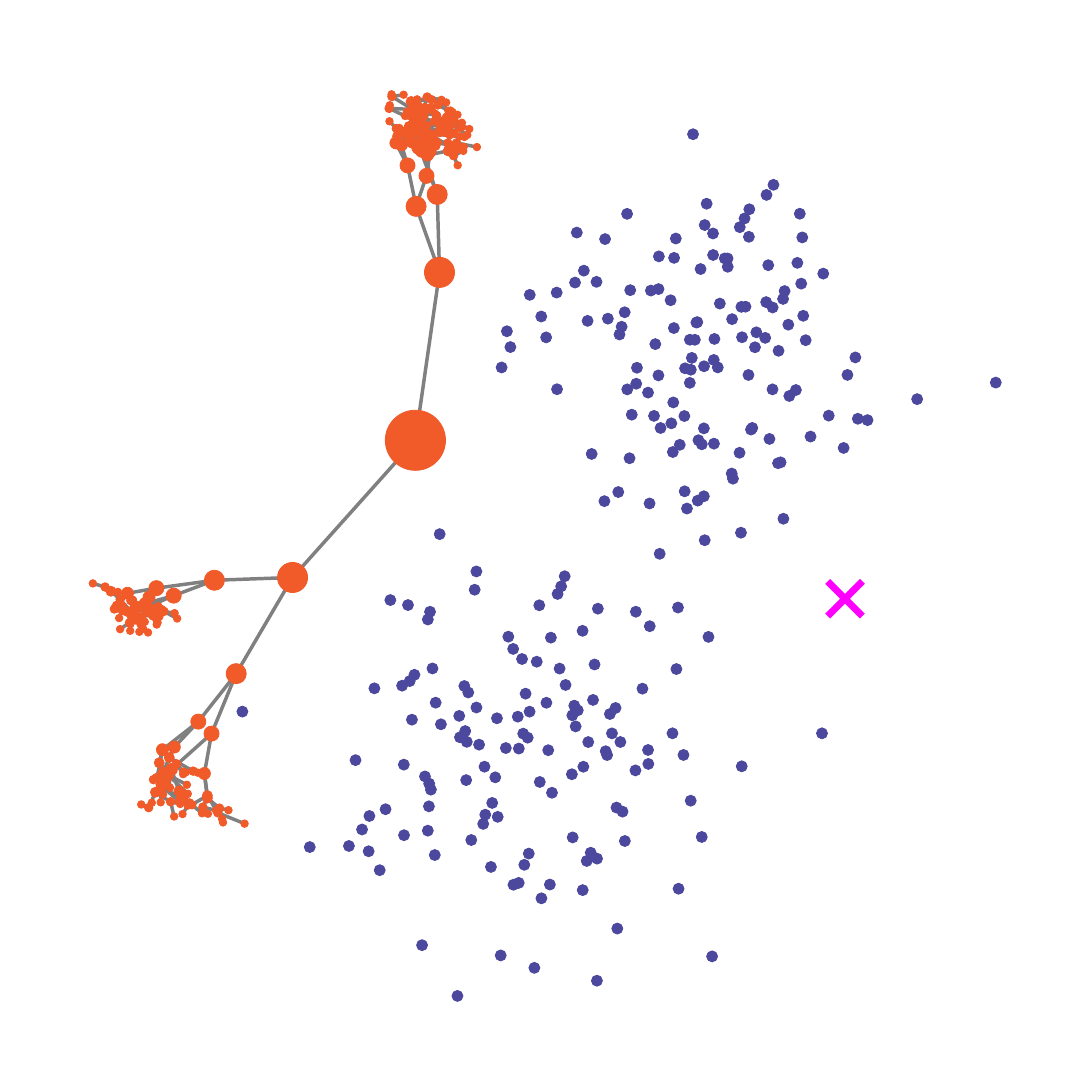}
      \label{fig:binary_tree_normal}
    \end{minipage}
    &
    \vspace{-0.25cm}
    \hspace{-0.25cm}
    \begin{minipage}[t]{0.47\hsize}
      \centering
      \subcaption{Hyperbolic VAE}
      \vspace{-0.25cm}
      \includegraphics[width=\linewidth]{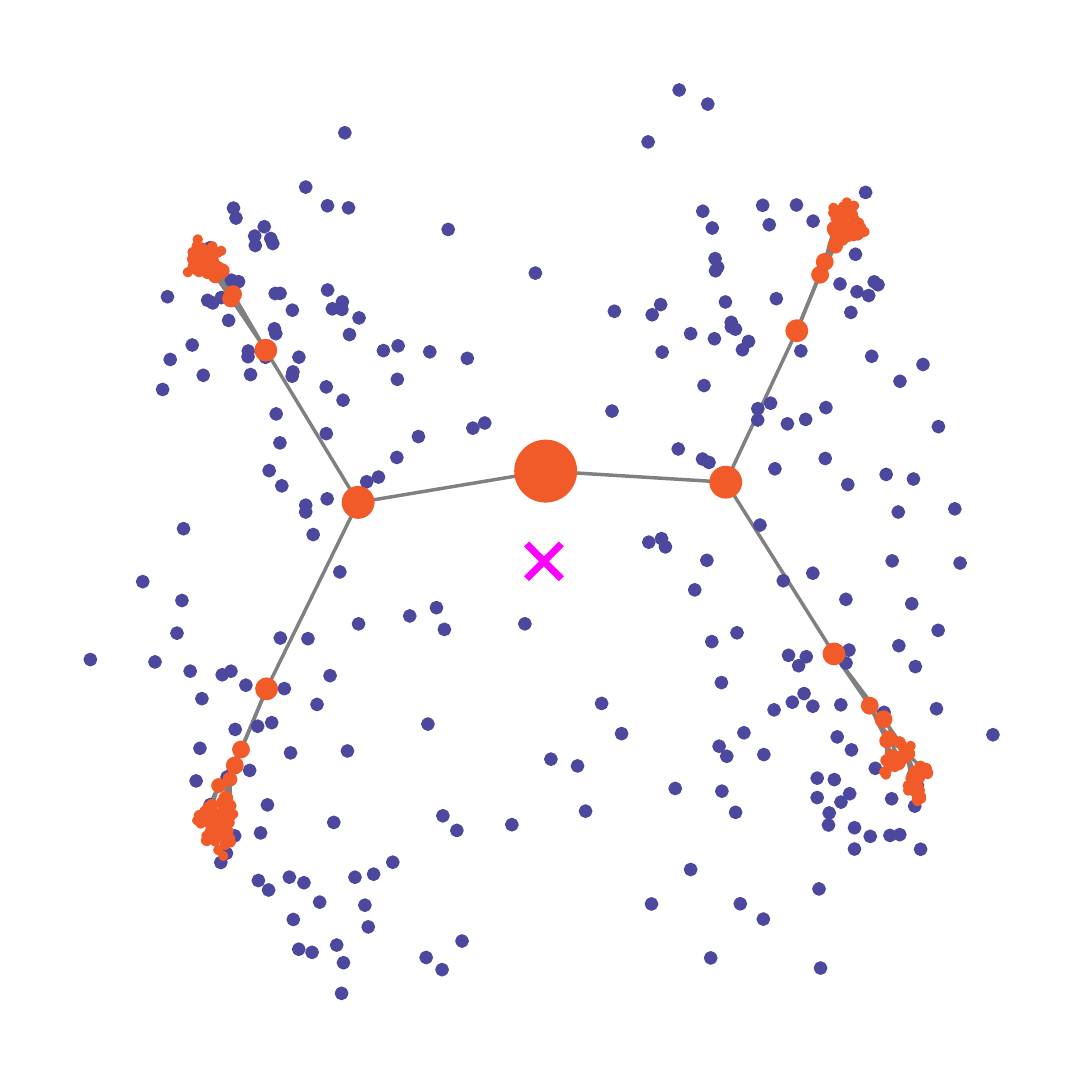}
      \label{fig:binary_tree_hyperbolic}
    \end{minipage}
  \end{tabular}
  \vspace{-0.35cm}
  \caption{The visual results of Hyperbolic VAE applied to an artificial dataset generated by applying random perturbations to a binary tree. The visualization is being done on the Poincar\'e ball.
  The red points are the embeddings of the original tree, and the blue points are the embeddings of noisy observations generated from the tree.
  The pink $\times$ represents the origin of the hyperbolic space.
  The VAE was trained without the prior knowledge of the tree structure.
  Please see \ref{sub:binary_tree} for experimental details
  }
  \label{fig:intro}
\end{figure}

In the task of data embedding, the choice of the target space determines the properties of the dataset that can be learned from the embedding.
For the dataset with a hierarchical structure, in particular, the number of relevant features can grow exponentially with the depth of the hierarchy.
Euclidean space is often inadequate for capturing the structural information (Figure \ref{fig:intro}).
If the choice of the target space of the embedding is limited to Euclidean space, one might have to prepare extremely high dimensional space as the target space to guarantee small distortion.
However, the same embedding can be done remarkably well if the destination is the hyperbolic space \cite{Sarkar2012,Sala2018}.

Now, the next natural question is; ``how can we extend these works to \textit{probabilistic} inference problems on hyperbolic space?''
When we know in advance that there is a hierarchical structure in the dataset, a prior distribution on hyperbolic space might serve as a good \textit{informative} prior.
We might also want to make Bayesian inference on a dataset with hierarchical structure by training a variational autoencoder (VAE)~\cite{Kingma2013,Rezende2014} with latent variables defined on hyperbolic space.
We might also want to conduct probabilistic word embedding into hyperbolic space while taking into account the uncertainty that arises from the underlying hierarchical relationship among words.
Finally, it would be best if we can compare different probabilistic models on hyperbolic space based on popular statistical measures like divergence that requires the explicit form of the probability density function.

The endeavors we mentioned in the previous paragraph all require \textit{probability distributions on hyperbolic space} that admit a parametrization of the density function that can be \textbf{computed analytically} and \textbf{differentiated} with respect to the parameter.
Also, we want to be able to \textbf{sample from the distribution efficiently}; that is, we do not want to resort to auxiliary methods like rejection sampling.

In this study, we present a novel hyperbolic distribution called \textit{pseudo-hyperbolic Gaussian}, a Gaussian-like distribution on hyperbolic space that resolves all these problems.
We construct this distribution by defining Gaussian distribution on the tangent space at the origin of the hyperbolic space and projecting the distribution onto hyperbolic space after transporting the tangent space to a desired location in the space.
This operation can be formalized by a combination of the parallel transport and the exponential map for the Lorentz model of hyperbolic space.

We can use our pseudo-hyperbolic Gaussian distribution to construct a probabilistic model on hyperbolic space that can be trained with gradient-based learning.
For example, our distribution can be used as a prior of a VAE (Figure \ref{fig:intro}, Figure \ref{fig:breakout_norm_controlled_samples}).
It is also possible to extend the existing probabilistic embedding method to hyperbolic space using our distribution, such as probabilistic word embedding.
We will demonstrate the utility of our method through the experiments of probabilistic hyperbolic models on benchmark datasets including MNIST, Atari 2600 Breakout, and WordNet.

\section{Background}

\begin{figure*}[htbp]
  \centering
  \begin{tabular}{ccc}
    \begin{minipage}[t]{0.3\hsize}
      \centering
      \subcaption{\leftline{}}
      \includegraphics[width=\linewidth]{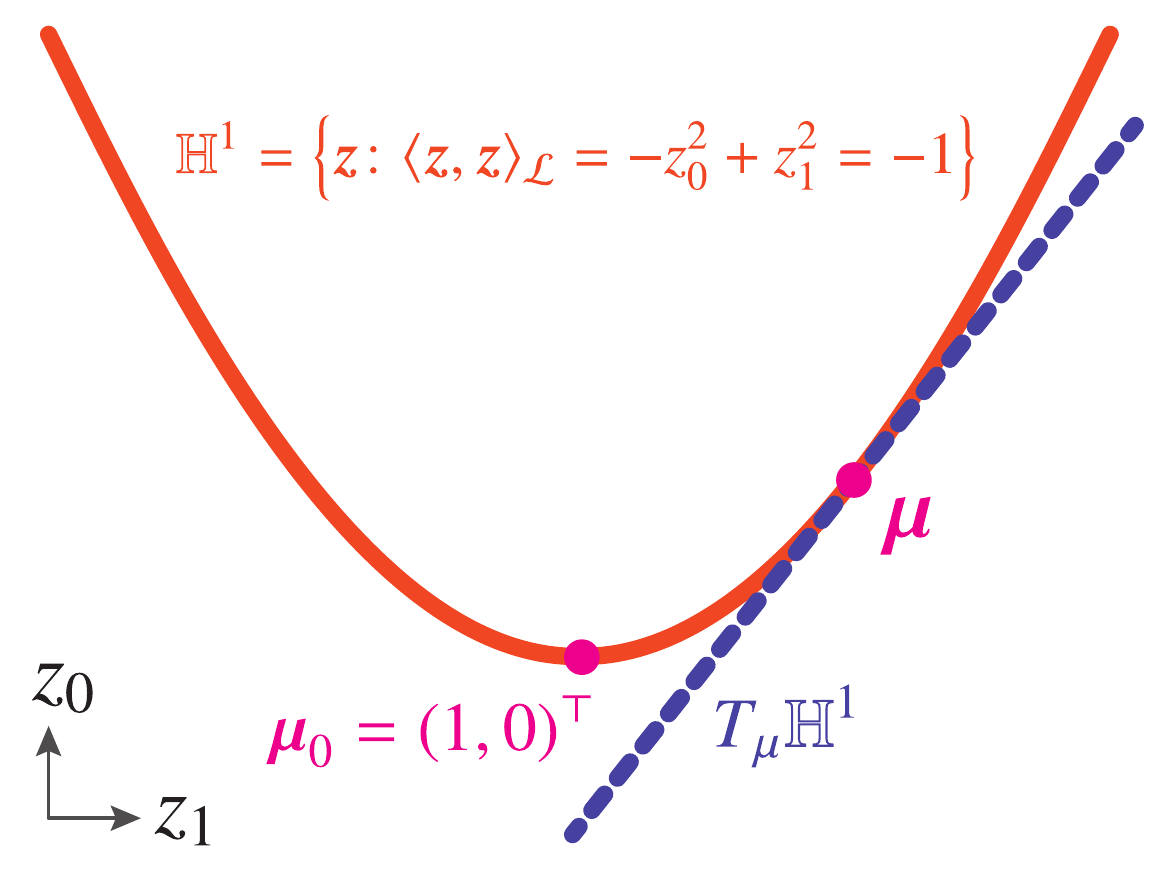}
      \label{fig:hyperboloid}
    \end{minipage}
    &
    \begin{minipage}[t]{0.3\hsize}
      \centering
      \subcaption{\leftline{}}
      \includegraphics[width=\linewidth]{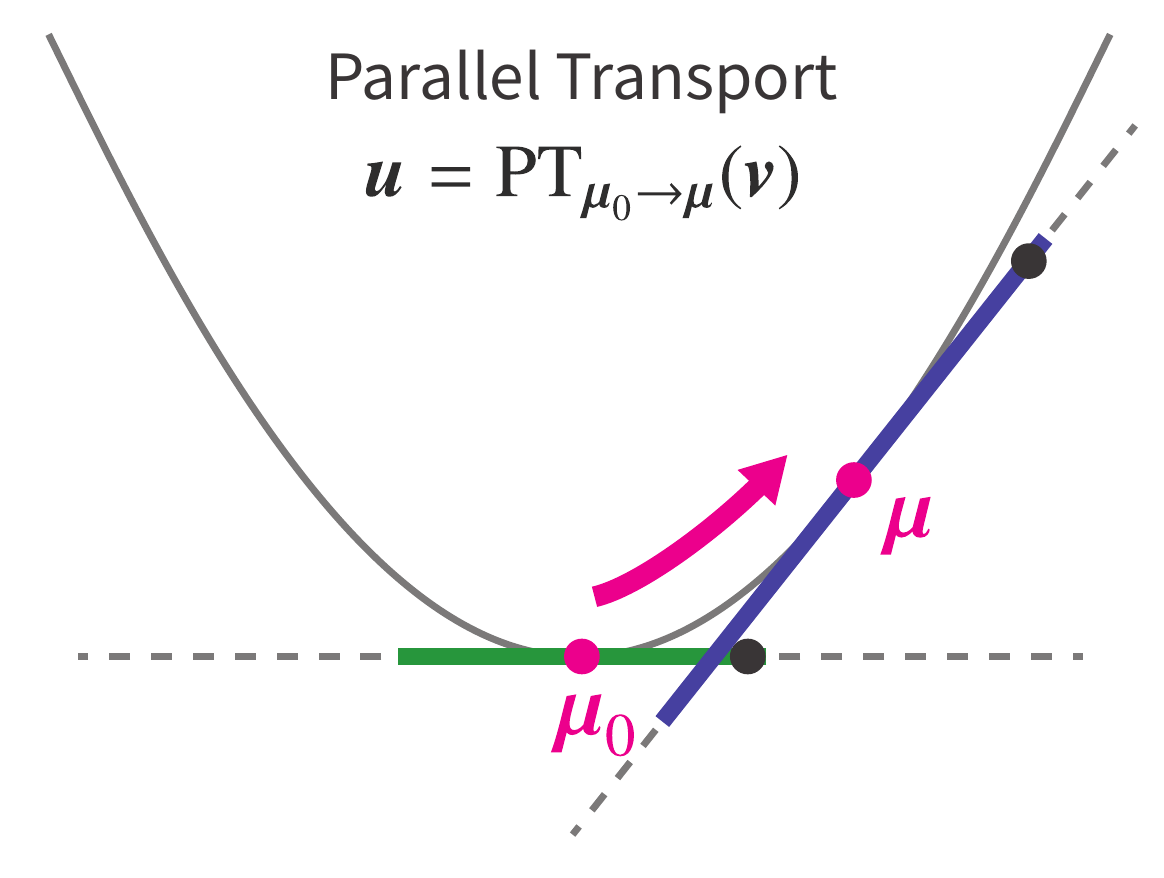}
      \label{fig:paramap}
    \end{minipage}
    &
    \begin{minipage}[t]{0.3\hsize}
      \centering
      \subcaption{\leftline{}}
      \includegraphics[width=\linewidth]{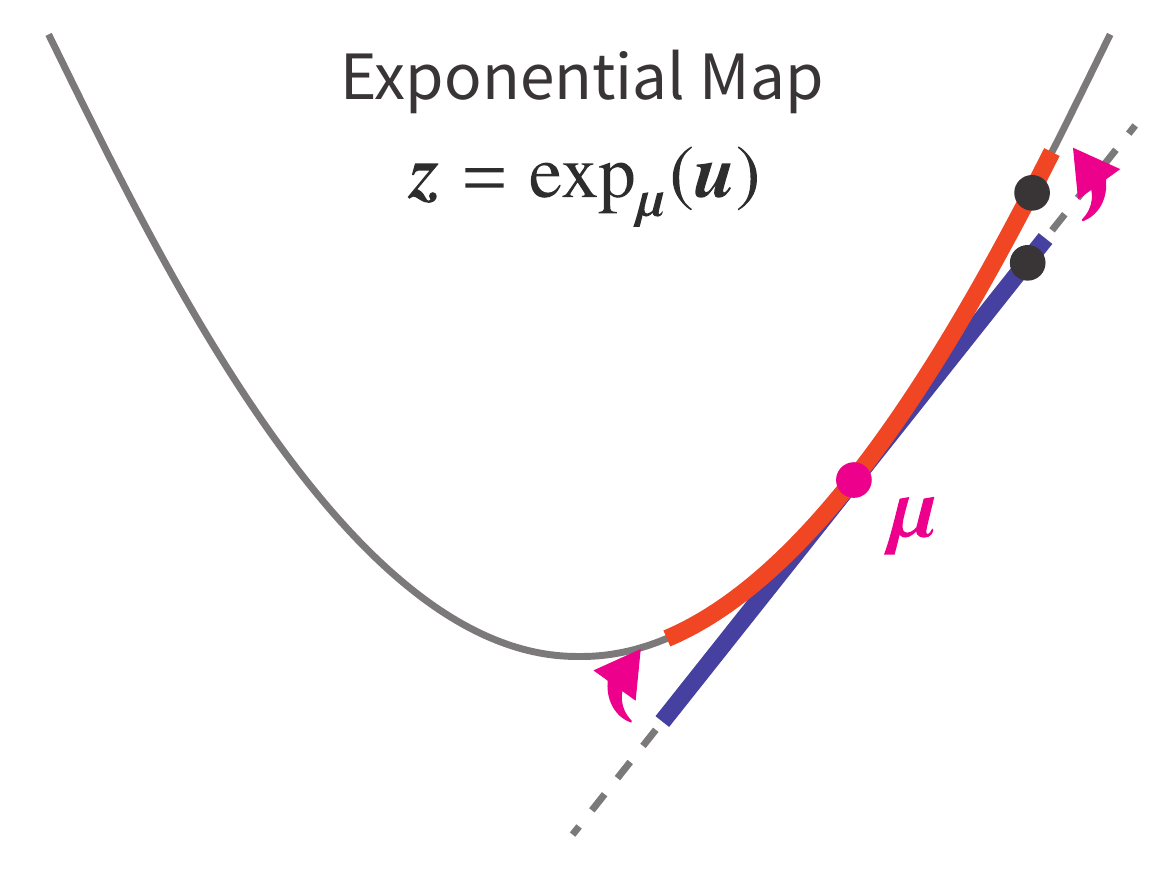}
      \label{fig:expmap}
    \end{minipage}

  \end{tabular}
  \vspace{-0.5cm}
  \caption{(a) One-dimensional Lorentz model $\mathbb{H}^1$ (red) and its tangent space $T_{\vb*{\mu}} \mathbb{H}^1$ (blue). (b) Parallel transport carries $\vb*{v} \in T_{\vb*{\mu}_0}$ (green) to $\vb*{u} \in T_{\vb*{\mu}}$ (blue) while preserving $\|\cdot \|_\mathcal{L}$ .
  (c) Exponential map projects the $\vb*{u} \in T_{\vb*{\mu}}$ (blue) to $\vb*{z} \in \mathbb{H}^n$ (red). The distance between $\vb*{\mu}$ and $\exp_{\vb*{\mu}}(\vb*{u})$ which is measured on the surface of $\mathbb{H}^n$ coincides with $\|\vb*{u}\|_\mathcal{L}$.}
  \label{fig:exp_para}
\end{figure*}

\subsection{Hyperbolic Geometry}
Hyperbolic geometry is a non-Euclidean geometry with a constant negative Gaussian curvature, and it can be visualized as the forward sheet of the two-sheeted hyperboloid.
There are four common equivalent models used for the hyperbolic geometry: the Klein model, Poincar\'e disk model, and Lorentz (hyperboloid/Minkowski) model, and Poincar\'e half-plane model.
Many applications of hyperbolic space to machine learning to date have adopted the Poincar\'e disk model as the subject of study \cite{Nickel2017,Ganea2018a,Ganea2018b,Sala2018}.
In this study, however, we will use the Lorentz model that, as claimed in
 \citet{Nickel2018}, comes with a simpler closed form of the geodesics and does not suffer from the numerical instabilities in approximating the distance.
We will also exploit the fact that both exponential map and parallel transportation have a clean closed form in the Lorentz model.

Lorentz model $\mathbb{H}^n$ (Figure \ref{fig:hyperboloid}) can be represented as a set of points $\vb*{z} \in \mathbb{R}^{n+1}$
with $z_0 > 0$ such that its Lorentzian product (negative Minkowski bilinear form)
\begin{equation*}
  \langle \vb*{z}, \vb*{z}' \rangle_{\mathcal{L}} = - z_0 z'_0 + \sum_{i=1}^n z_i z'_i,
\end{equation*}
with itself is $-1$. That is,
\begin{equation}
  \mathbb{H}^n = \qty{\vb*{z} \in \mathbb{R}^{n+1} \colon \langle \vb*{z}, \vb*{z} \rangle_{\mathcal{L}} = -1, \:\: z_0 > 0}. \label{eq:lorentz_def}
\end{equation}
Lorentzian inner product also functions as the metric tensor on hyperbolic space.
We will refer to the one-hot vector $\vb*{\mu}_0 = [1, 0,0,...0] \in \mathbb{H}^n \subset \mathbb{R}^{n+1}$ as the \textit{origin} of the  hyperbolic space.
Also, the distance between two points $\vb*{z}, \vb*{z}'$ on $\mathbb{H}^n$ is given by
$d_{\ell}(\vb*{z},\vb*{z}') = \arccosh \qty( - \langle \vb*{z}, \vb*{z}' \rangle_{\mathcal{L}} )$, which is also the length of the geodesic that connects $\bm{z}$ and $\bm{z}'$.

\subsection{Parallel Transport and Exponential Map}
The rough explanation of our strategy for the construction of pseudo-hyperbolic Gaussian $\mathcal{G}(\vb*{\mu}, \Sigma)$ with $\vb*{\mu} \in \mathbb{H}^n$ and a positive positive definite matrix $\Sigma$ is as follows.
We (1) sample a vector from $\mathcal{N}(\vb*{0}, \Sigma)$, (2) transport the vector from $\vb*{\mu}_0$ to $\vb*{\mu}$ along the geodesic, and (3) project the vector onto the surface.
To formalize this sequence of operations, we need to define the tangent space on hyperbolic space as well as the way to transport the tangent space and the way to project a vector in the tangent space to the surface.
The transportation of the tangent vector requires \textit{parallel transport}, and the projection of the tangent vector to the surface requires the definition of \textit{exponential map.}

{\bf Tangent space of hyperbolic space}

Let us use $T_{\bm{\mu}}\mathbb{H}^n$ to denote the tangent space of $\mathbb{H}^n$ at $\bm{\mu}$ (Figure \ref{fig:hyperboloid}).
Representing $T_{\bm{\mu}}\mathbb{H}^n$ as a set of vectors in the same ambient space $\mathbb{R}^{n+1}$ into which $\mathbb{H}^n$ is embedded, $T_{\bm{\mu}}\mathbb{H}^n$ can be characterized as the set of points satisfying the orthogonality relation with respect to the Lorentzian product:
\begin{align}
    T_{\bm{\mu}}\mathbb{H}^n := \{\bm{u} \colon \langle \bm{u}, \bm{\mu} \rangle_{\mathcal{L}} = 0 \}.
    \label{eq:tangent}
\end{align}
$T_{\bm{\mu}}\mathbb{H}^n$ set can be literally thought of as the tangent space of the forward hyperboloid sheet at $\vb*{\mu}$.
Note that $T_{\vb*{\mu}_0}\mathbb{H}^n$ consists of $\bm{v} \in \mathbb{R}^{n+1}$ with $\vb*{v}_0 = 0$, and $\|\bm{v}\|_\mathcal{L} := \sqrt{\langle \vb*{v}, \vb*{v} \rangle_{\mathcal{L}}} =  \|\bm{v}\|_2$.

{\bf Parallel transport and inverse parallel transport}\\
Next, for an arbitrary pair of point $\bm{\mu}, \bm{\nu} \in \mathbb{H}^{n}$, the parallel transport from $\vb*{\nu}$ to $\vb*{\mu}$ is defined as a map $\PT_{\bm{\nu} \to \bm{\mu}}$ from $T_{\bm{\nu}}\mathbb{H}^n$ to $T_{\bm{\mu}}\mathbb{H}^n$ that carries a vector in $T_{\bm{\nu}}\mathbb{H}^n$ along the geodesic from $\vb*{\nu}$ to $\vb*{\mu}$ in a parallel manner without changing its metric tensor.
In other words, if $\PT$ is the parallel transport on hyperbolic space, then $\langle \PT_{\bm{\nu} \to \bm{\mu}}(\bm{v}), \PT_{\bm{\nu} \to \bm{\mu}}(\bm{v}') \rangle_\mathcal{L} = \langle \bm{v}, \vb*{v}' \rangle_\mathcal{L}$.

The explicit formula for the parallel transport on the Lorentz model (Figure \ref{fig:paramap}) is given by:

\begin{align}
  \PT_{\vb*{\nu} \rightarrow \vb*{\mu}} & (\vb*{v}) = \vb*{v} + \frac{ \langle \vb*{\mu} - \alpha \vb*{\nu}, \vb*{v} \rangle_{\mathcal{L}} }{\alpha + 1} \qty( \vb*{\nu} + \vb*{\mu} ), \label{eq:parallel}
\end{align}
\noindent
where $\alpha = - \langle \vb*{\nu}, \vb*{\mu} \rangle_{\mathcal{L}}$.
The inverse parallel transport $\PT_{\vb*{\nu} \rightarrow \vb*{\mu}}^{-1}$ simply carries the vector in $T_{\bm{\mu}}\mathbb{H}^n$ back to $T_{\bm{\nu}}\mathbb{H}^n$ along the geodesic. That is,

\begin{equation}
  \vb*{v} = \PT_{\vb*{\nu} \rightarrow \vb*{\mu}}^{-1} (\vb*{u}) = \PT_{\vb*{\mu} \rightarrow \vb*{\nu}} (\vb*{u}). \label{eq:inv_parallel}
\end{equation}

{\bf Exponential map and inverse exponential map} \\
Finally, we will describe a function that maps a vector in a tangent space to its surface.

According to the basic theory of differential geometry, every $\vb*{u} \in T_{\bm{\mu}}\mathbb{H}^n $ determines a unique maximal geodesic $\gamma_{\bm{\mu}}:[0,1] \to \mathbb{H}^n$ with $\gamma_{\bm{\mu}}(0)= \vb*{\mu}$ and $\dot \gamma_{\bm{\mu}}(0) = \vb*{u}$. 
Exponential map $\exp_{\bm{\mu}}: T_{\bm{\mu}}\mathbb{H}^n \to \mathbb{H}^n$ is a map defined by $\exp_{\bm{\mu}}(\vb*{u}) = \gamma_{\bm{\mu}}(1)$,
and we can use this map to project a vector $\vb*{v}$ in $T_{\bm{\mu}}\mathbb{H}^n$ onto $\mathbb{H}^n$ in a way that the distance from $\vb*{\mu}$ to destination of the map coincides with $\|\vb*{v}\|_{\mathcal{L}}$, the metric norm of $\vb*{v}$.
For hyperbolic space, this map (Figure \ref{fig:expmap}) is given by

\begin{equation}
  \vb*{z} = \exp_{\vb*{\mu}}(\vb*{u}) = \cosh \qty( \norm{\vb*{u}}_{\mathcal{L}} ) \vb*{\mu} + \sinh \qty( \norm{\vb*{u}}_{\mathcal{L}} ) \frac{\vb*{u}}{\norm{\vb*{u}}_{\mathcal{L}}}. \label{eq:expmap}
\end{equation}

\noindent
As we can confirm with straightforward computation, this exponential map is norm preserving in the sense that
$d_{\ell}(\vb*{\mu}, \exp_{\bm{\mu}}(\vb*{u})) = \arccosh \qty( - \langle \vb*{\mu}, \exp_{\bm{\mu}}(\vb*{u}) \rangle_{\mathcal{L}} ) = \| \vb*{u} \|_{\mathcal{L}}$.
Now, in order to evaluate the density of a point on hyperbolic space, we need to be able to map the point back to the tangent space, on which the distribution is initially defined.
We, therefore, need to be able to compute the inverse of the exponential map, which is also called logarithm map, as well.

Solving  \cref{eq:expmap} for $\vb*{u}$, we can obtain the inverse exponential map as
\begin{equation}
  \vb*{u} = \exp^{-1}_{\vb*{\mu}}(\vb*{z}) = \frac{ \arccosh(\alpha) }{ \sqrt{\alpha^2 - 1} } \qty(\vb*{z} - \alpha \vb*{\mu}), \label{eq:inv_expmap}
\end{equation}
where $\alpha = - \langle \vb*{\mu}, \vb*{z} \rangle_{\mathcal{L}}$.
See Appendix A.1 for further details.

\section{Pseudo-Hyperbolic Gaussian}
\label{sec:prob_on_hyperbola}

\subsection{Construction}
Finally, we are ready to provide the construction of our wrapped gaussian distribution $\mathcal{G}(\vb*{\mu}, \Sigma)$ on Hyperbolic space with $\vb*{\mu} \in \mathbb{H}^n$ and positive definite $\Sigma$.

In the language of the differential geometry,
our strategy can be re-described as follows:
\begin{enumerate}
    \item Sample a vector $\bm{\tilde{v}}$ from the  Gaussian distribution $\mathcal{N}(\vb*{0}, \Sigma)$ defined over $\mathbb{R}^n$.
    \item Interpret $\bm{\tilde{v}}$ as an element of $T_{\bm{\mu}_0}\mathbb{H}^n \subset \mathbb{R}^{n+1}$ by rewriting  $\bm{\tilde{v}}$ as
    $\vb*{v} = [0, \vb*{\tilde{v}}]$.
    \item Parallel transport the vector $\bm{v}$ to $\bm{u} \in T_{\bm{\mu}}\mathbb{H}^n \subset \mathbb{R}^{n+1}$ along the geodesic from $\bm{\mu}_0$ to $\bm{\mu}$.
    \item Map $\bm{u}$ to $\mathbb{H}^n$ by $\exp_{\bm{\mu}}$.
\end{enumerate}

Algorithm \ref{alg:sampling} is an algorithmic description of the sampling procedure based on our construction.

\begin{algorithm}[t]
  \caption{Sampling on hyperbolic space}
  \label{alg:sampling}
  \begin{algorithmic}
    \STATE {\bfseries Input:} parameter $\vb*{\mu} \in \mathbb{H}^n$, $\Sigma$
    \STATE {\bfseries Output:} $\vb*{z} \in \mathbb{H}^n$
    \STATE {\bfseries Require:} $\vb*{\mu}_{\mathrm{0}} = (1, 0, \cdots, 0)^\top \in \mathbb{H}^n$
    \STATE Sample $\tilde{\vb*{v}} \sim \mathcal{N}(\vb*{0}, \Sigma) \in \mathbb{R}^n$
    \STATE $\vb*{v} = [0, \tilde{\vb*{v}}] \in T_{\vb*{\mu}_{\mathrm{0}}} \mathbb{H}^n$
    \STATE Move $\vb*{v}$ to $\vb*{u} = \PT_{\vb*{\mu}_{\mathrm{0}} \rightarrow \vb*{\mu}} (\vb*{v}) \in T_{\vb*{\mu}} \mathbb{H}^n$ by \cref{eq:parallel}
    \STATE Map $\vb*{u}$ to $\vb*{z} = \exp_{\vb*{\mu}}(\vb*{u}) \in \mathbb{H}^n$ by \cref{eq:expmap}
  \end{algorithmic}
\end{algorithm}
The most prominent advantage of this construction is that we can compute the density of the probability distribution.

\subsection{Probability Density Function}

Note that both $\PT_{\bm{\mu}_0 \to \bm{\mu}}$ and $\exp_{\bm{\mu}}$ are differentiable functions that can be evaluated analytically.
Thus, by the construction of $\mathcal{G}(\bm{\mu}, \Sigma)$, we can compute the probability density of $\mathcal{G}(\bm{\mu}, \Sigma)$ at $\bm{z} \in \mathbb{H}^n$ using a composition of differentiable functions,
$\PT_{\bm{\mu}_0 \to \bm{\mu}}$ and $\exp_{\bm{\mu}}$.
Let $\proj_{\bm{\mu}}  := \exp_{\bm{\mu}} \circ  \PT_{\bm{\mu}_0 \to \bm{\mu}} $ (Figure \ref{sec:prob_on_hyperbola}).

In general, if $\bm{X}$ is a random variable endowed with the probability density function $p(\bm{x})$, the log likelihood of $Y = f(\bm{X})$ at $\bm{y}$ can be expressed as
\begin{align*}
    \log p(\bm{y}) =  \log p(\bm{x}) - \log \det \left(\frac{\partial f}{\partial \bm{x} }  \right)
\end{align*}
where $f$ is a invertible and continuous map. Thus, all we need in order to evaluate the probability density of $\mathcal{G}(\bm{\mu}, \Sigma)$ at $\bm{z} = \proj_{\vb*{\mu}}(\bm{v})$ is the way to evaluate $\det (\partial \proj_{\vb*{\mu}}(\bm{v}) / \partial \bm{v})$:
\begin{equation}
  \log p(\vb*{z}) = \log p(\vb*{v}) - \log \det \left(\pdv{\proj_{\vb*{\mu}}(\vb*{v})}{\vb*{v}} \right). \\
  \label{eq:logpdf}
\end{equation}
Algorithm \ref{alg:logpdf} is an algorithmic description for the computation of the pdf.

\begin{algorithm}[t]
  \caption{Calculate log-pdf}
  \label{alg:logpdf}
  \begin{algorithmic}
    \STATE {\bfseries Input:} sample $\vb*{z} \in \mathbb{H}^n$, parameter $\vb*{\mu} \in \mathbb{H}^n$, $\Sigma$
    \STATE {\bfseries Output:} $\log p(\vb*{z})$
    \STATE {\bfseries Require:} $\vb*{\mu}_{\mathrm{0}} = (1, 0, \cdots, 0)^\top \in \mathbb{H}^n$
    \STATE Map $\vb*{z}$ to $\vb*{u} = \exp^{-1}_{\vb*{\mu}}(\vb*{z}) \in T_{\vb*{\mu}} \mathbb{H}^n$ by \cref{eq:inv_expmap}
    \STATE Move $\vb*{u}$ to $\vb*{v} = \PT^{-1}_{\vb*{\mu}_{\mathrm{0}} \rightarrow \vb*{\mu}}(\vb*{u}) \in T_{\vb*{\mu}_{\mathrm{0}}} \mathbb{H}^n$ by \cref{eq:inv_parallel}
    \STATE Calculate $\log p(\vb*{z})$ by \cref{eq:logpdf}
  \end{algorithmic}
\end{algorithm}

\noindent
For the implementation of \cref{alg:sampling} and \cref{alg:logpdf}, we would need to be able to evaluate not only $\exp_{\vb*{\mu}}(\vb*{u})$,  $\PT_{\vb*{\mu}_{\mathrm{0}} \rightarrow \vb*{\mu}}(\vb*{v})$ and their inverses, but also need to evaluate the determinant.
We provide an analytic solution to each one of them below.

{\bf Log-determinant}\\
For the evaluation of \eqref{eq:logpdf}, we need to compute the
log determinant of the projection function that maps a vector in the tangent space $T_{\mu_0}(\mathbb{H}^n)$ at origin to the tangent space $T_{\mu}(\mathbb{H}^n)$ at an arbitrary point $\vb*{\mu}$ in the hyperbolic space.

Appealing to the chain-rule and the property of determinant, we can decompose the expression into two components:
\begin{align}
  \det & \qty( \pdv{\proj_{\vb*{\mu}}(\vb*{v})}{\vb*{v}} ) \nonumber \\
  &= \det \qty( \pdv{\exp_{\vb*{\mu}}(\vb*{u})}{\vb*{u}}) \cdot \det \qty( \pdv{ \PT_{ \vb*{\mu}_0 \rightarrow \vb*{\mu}}(\vb*{v}) }{ \vb*{v} } ). \label{eq:det}
\end{align}
\noindent
We evaluate each piece one by one.
First, let us recall that $\partial \exp_{\vb*{\mu}}(\vb*{u}) / \partial \vb*{u}$ is a map that sends an element in $T_u(T_\mu(\mathbb{H}^{n})) =T_\mu(\mathbb{H}^{n})$ to an element in $T_v(\mathbb{H}^{n})$, where $\vb*{v} = \exp_{\vb*{\mu}}(\vb*{u})$.
We have a freedom in choosing a basis to evaluate the determinant of this expression.
For convenience, let us choose an orthonormal basis of $T_\mu(\mathbb{H}^{n})$ that contains $\bar{\vb*{u}} = \vb*{u} / \norm{\vb*{u}}_{\mathcal{L}}$:
$$\{\bar{\vb*{u}}, \vb*{u}^{\prime}_1, \vb*{u}^{\prime}_2, ....\vb*{u}^{\prime}_{n-1} \}$$

Now, computing the directional derivative of $\exp_\mu$ with respect to each basis element, we get
\begin{align}
  \dd \exp_{\vb*{\mu}}(\bar{\vb*{u}}) =& \sinh(r) \vb*{\mu} + \cosh(r) \bar{\vb*{u}}, \\
  \dd \exp_{\vb*{\mu}}(\vb*{u}^\prime_k) =& \frac{\sinh r}{r} \vb*{u}^\prime_k,
\end{align}
where $r=\norm{\vb*{u}}_{\mathcal{L}}$.
\noindent
Because the directional derivative in the direction of $\bar{\vb*{u}}$ has magnitude $1$ and because each components are orthogonal to each other, the norm of the directional derivative is given by
\begin{equation}
  \det \qty( \pdv{\exp_{\vb*{\mu}}(\vb*{u})}{\vb*{u}}) = \qty(\frac{\sinh r}{r})^{n-1}.
\end{equation}
which yields the value of the desired determinant.
\noindent
Let us next compute the determinant of the derivative of parallel transport.
Evaluating the determinant with an orthogonal basis of $T_v(T_{\mu_0}(\mathbb{H}^n))$ , we get
$$\dd \PT_{\vb*{\mu}_0 \rightarrow \vb*{\mu}}(\vb*{\xi}) = \PT_{\vb*{\mu}_0 \rightarrow \vb*{\mu}}(\vb*{\xi}).$$
Because parallel transport is itself a norm preserving map, this means that $\det (\partial \PT_{ \vb*{\mu}_0 \rightarrow \vb*{\mu}}(\vb*{v}) / \partial \vb*{v} ) = 1$.

All together, we get
\begin{align}
  \det & \qty( \pdv{\proj_{\vb*{\mu}}(\vb*{v})}{\vb*{v}} ) = \qty( \frac{\sinh r}{r} )^{n-1}. \nonumber
\end{align}
We will provide the details of the computation in Appendix A.3. and A.4.
\noindent
The whole evaluation of the log determinant can be computed in $\mathcal{O}(n)$.

Since the metric at the tangent space coincides with the  Euclidean metric, we can produce various types of Hyperbolic distributions by applying our construction strategy to other distributions defined on Euclidean space, such as Laplace and Cauchy distribution.

\begin{figure}[!t]
  \centering
  \begin{minipage}[t]{\hsize}
      \centering
      \subcaption{\leftline{}}
      \vspace{-0.1cm}
      \includegraphics[width=\linewidth]{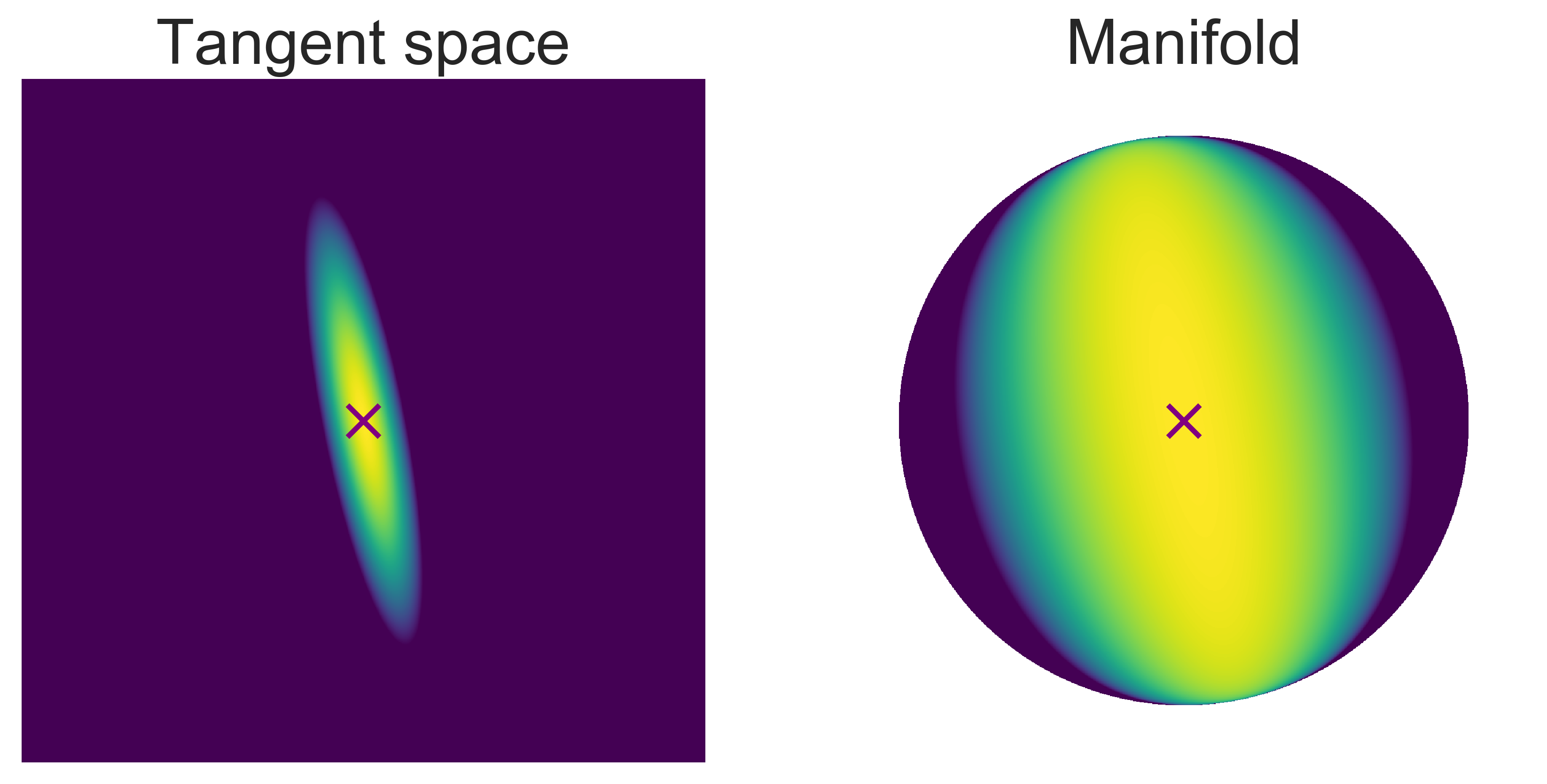}
    \end{minipage}\\
    \vspace{0.2cm}
    \begin{minipage}[t]{\hsize}
      \centering
      \subcaption{\leftline{}}
      \vspace{-0.1cm}
      \includegraphics[width=\linewidth]{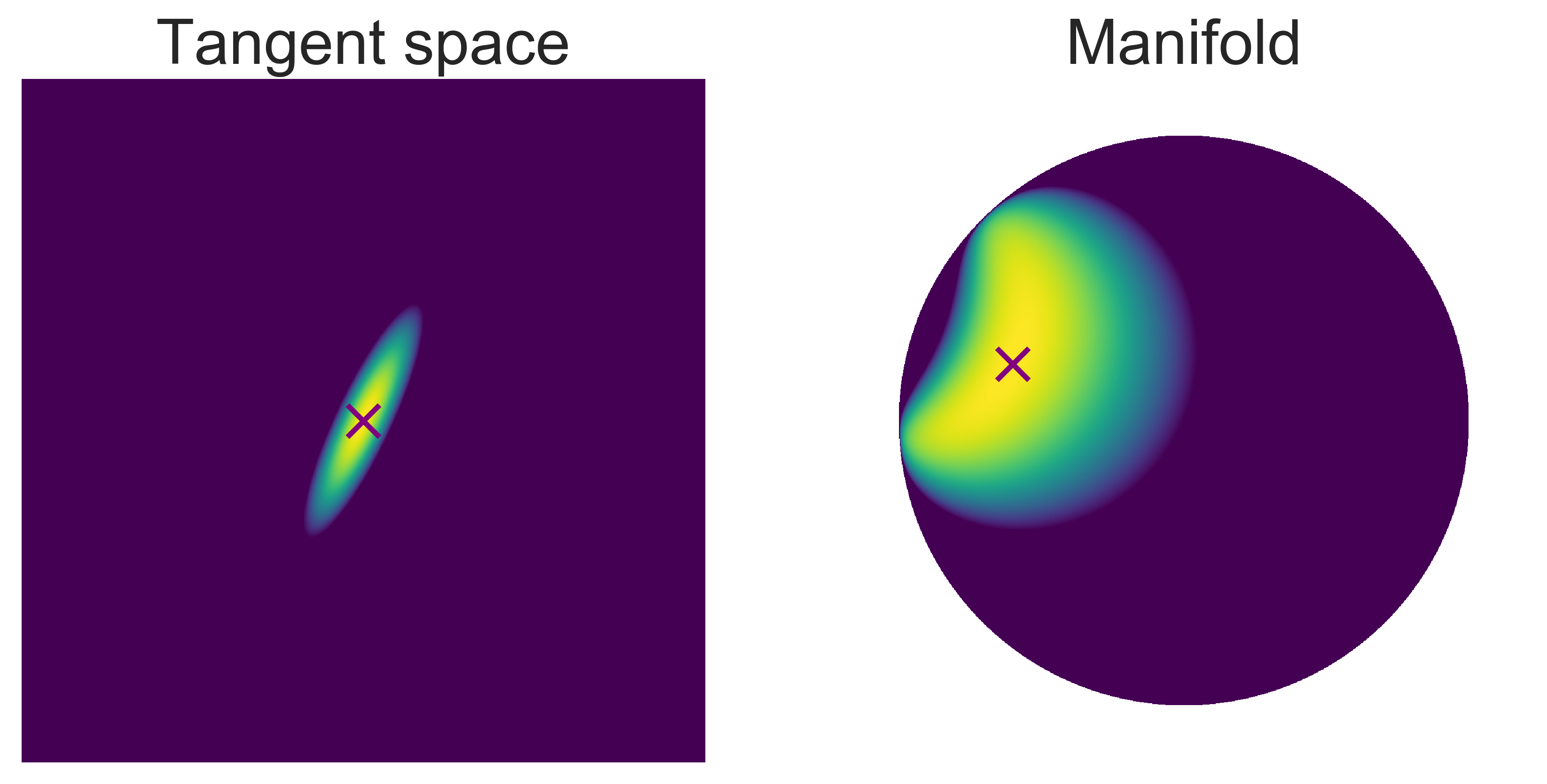}
    \end{minipage}
  \caption{The heatmaps of log-likelihood of the pesudo-hyperbolic Gaussians with various $\vb*{\mu}$ and $\Sigma$. We designate the origin of hyperbolic space by the $\times$ mark.
  See Appendix B for further details.
  }
  \label{fig:density}
\end{figure}

\section{Applications of $\mathcal{G}(\vb*{\mu}, \Sigma)$}

\subsection{Hyperbolic Variational Autoencoder}
As an application of pseudo-hyperbolic Gaussian $\mathcal{G}(\vb*{\mu}, \Sigma)$,
we will introduce \textit{hyperbolic variational autoencoder} (Hyperbolic VAE), a variant of the variational autoencoder (VAE) \cite{Kingma2013,Rezende2014} in which the latent variables are defined on hyperbolic space.
Given dataset $\mathcal{D} = \{\bm{x}^{(i)}\}_{i=1}^N$,
the method of variational autoencoder aims to train a decoder model $p_{\bm{\theta}}(\vb*{x} | \bm{z})$ that can create from $p_{\bm{\theta}}(\bm{z})$ a dataset that resembles $\mathcal{D}$.
The decoder model is trained together with the encoder model $q_{\bm{\phi}}(\bm{z} | \vb*{x})$ by maximizing the sum of evidence lower bound (ELBO) that is defined for each $\bm{x}^{(i)}$;
\begin{align}
  & \log p_{\vb*{\theta}}(\vb*{x}^{(i)}) \ge \mathcal{L}(\vb*{\theta}, \vb*{\phi}; \vb*{x}^{(i)}) = \nonumber \\
  & \mathbb{E}_{q_{\vb*{\phi}}(\vb*{z}|\vb*{x}^{(i)})} \qty[ \log p_{\vb*{\theta}} (\vb*{x}^{(i)}|\vb*{z}) ] - D_{\mathrm{KL}} (q_{\vb*{\phi}}(\vb*{z}|\vb*{x}^{(i)})||p_{\vb*{\theta}} (\vb*{z})), \label{eq:elbo}
\end{align}
where $q_{\bm{\phi}}(\bm{z}| \vb*{x}^{(i)})$ is the variational posterior distribution.
In classic VAE, the choice of the prior $p_{\bm{\theta}}$ is the standard normal, and the posterior distribution is also variationally approximated by a Gaussian.
Hyperbolic VAE is a simple modification of the classic VAE in which $p_{\bm{\theta}} = \mathcal{G}(\bm{\mu}_0, I)$ and $q_{\bm{\phi}} = \mathcal{G}(\bm{\mu}, \Sigma)$.
The model of $\bm{\mu}$ and $\Sigma$ is often referred to as encoder.
This parametric formulation of $q_{\bm{\phi}}$ is called reparametrization trick, and it enables the evaluation of the gradient of the objective function with respect to the network parameters.
To compare our method against,  we used $\beta$-VAE~\cite{Higgins2017}, a variant of VAE that applies a scalar weight $\beta$ to the KL term in the objective function.

In Hyperbolic VAE, we assure that output $\mu$ of the encoder is in  $\in \mathbb{H}^n$ by applying $\exp_{\bm{\mu}_0}$ to the final layer of the encoder.
That is,  if $\bm{h}$ is the output, we can simply use
\begin{equation*}
  \vb*{\mu} = \exp_{\bm{\mu}_0}(\bm{h}) =  \mqty( \cosh(\norm{\vb*{h}}_2) ,& \sinh(\norm{\vb*{h}}_2) \frac{ \vb*{h} }{ \norm{\vb*{h}}_2 } )^\top.
\end{equation*}
As stated in the previous sections, our distribution $\mathcal{G}(\bm{\mu}, \Sigma)$ allows us to evaluate the ELBO exactly and to take the gradient of the objective function.
In a way, our distribution of the variational posterior is an hyperbolic-analog of the reparametrization trick.

\subsection{Word Embedding}
We can use our psudo-hyperbolic Gaussian $\mathcal{G}$ for probabilistic word embedding.
The work of \citet{Vilnis2015} attempted to extract the linguistic and contextual properties of words in a dictionary by embedding every word and every context to a Gaussian distribution defined on Euclidean space.
We may extend their work by changing the destination of the map to the family of $\mathcal{G}$.
Let us write $a \sim b$ to convey that there is a link between words $a$ and $b$, and let us use $q_s$ to designate the distribution to be assigned to the word $s$.
The objective function used in \citet{Vilnis2015} is given by
\begin{equation*}
  \mathcal{L}(\vb*{\theta}) = \mathbb{E}_{(s \sim t, s \not \sim t')} \qty[ \max \qty( 0, m + E(s, t) - E(s, t') ) ],
\end{equation*}
where $E(s,t)$ represents the measure of similarity between $s$ and $t$ evaluated with
$D_{\mathrm{KL}} ( q_s \| q_t )$.
In the original work, $q_s$ and $q_t$ were chosen to be a Gaussian distribution.
We can incorporate hyperbolic geometry into this idea by choosing $q_s = \mathcal{G}(\vb*{\mu}(s), \Sigma(s))$.

\section{Related Work}
\label{sec:related}
The construction of hyperbolic distribution based on projection is not entirely new on its own.
For example, CCM-AAE \cite{Grattarola2018} uses a prior distribution on hyperbolic space centered at origin by projecting a distribution constructed on the tangent space.
Wrapped normal distribution (on sphere) also is a creation of similar philosophy.
Still yet, as mentioned in the introduction, most studies to date that use hyperbolic space consider only deterministic mappings \cite{Nickel2017,Nickel2018,Ganea2018a,Ganea2018b,Gulcehre2018}.

Very recently, \citet{Ovinnikov2019} proposed an application of Gaussian distribution on hyperbolic space.
However, the formulation of their distribution cannot be directly differentiated nor evaluated because of the presence of \textit{error function} in their expression of pdf.
For this reason, they resort to Wasserstein Maximum Mean Discrepancy \cite{Gretton2012} to train their encoder network.
Our distribution $\mathcal{G}$ has broader application than the distribution of \citet{Ovinnikov2019} because it allows the user to compute its likelihood and its gradient without approximation.
One advantage of our distribution $\mathcal{G}(\vb*{\mu}, \Sigma)$ is its representation power.
Our distribution $\mathcal{G}(\vb*{\mu}, \Sigma)$ can be defined for any $\vb*{\mu}$ in $\mathbb{H}^n$ and any positive definite matrix $\Sigma \in \mathbb{R}^{n \times n}$.
Meanwhile, the hyperbolic Gaussian studied in \citet{Ovinnikov2019} can only express Gaussian with variance matrix of the form $\sigma I$.

For word embedding, several deterministic methods have been proposed to date, including the celebrated Word2Vec \cite{Mikolov2013}. The aforementioned \citet{Nickel2017} uses deterministic hyperbolic embedding to exploit the hierarchical relationships among words.
The probabilistic word embedding was first proposed by \citet{Vilnis2015}.
As stated in the method section, their method maps each word to a Gaussian distribution on Euclidean space.
Their work suggests the importance of investigating the uncertainty of word embedding.
In the field of representation learning of word vectors, our work is the first in using hyperbolic probability distribution for word embedding.

On the other hand, the idea to use a noninformative, non-Gaussian prior in VAE is not new.
For example, \citet{Davidson2018} proposes the use of von Mises-Fisher prior, and \citet{Rolfe2017,Jang2017}
use discrete distributions as their prior.
With the method of Normalizing flow \cite{Rezende2015}, one can construct even more complex priors as well \cite{Kingma2016}.
The appropriate choice of the prior shall depend on the type of dataset.
As we will show in the experiment section, our distribution is well suited to the dataset with underlying tree structures.
Another choice of the VAE prior that specializes in such dataset has been proposed by \citet{Vikram2018}．
For the sampling, they use time-marginalized coalescent, a model that samples a random tree structure by a stochastic process.
Theoretically, their method can be used in combination with our approach by replacing their Gaussian random walk with a hyperbolic random walk.
\vspace{-0.1cm}
\section{Experiments}
\label{sec:experiments}

\subsection{Synthetic Binary Tree}
\label{sub:binary_tree}
We trained Hyperbolic VAE for an artificial dataset constructed from a binary tree of depth $d =8$.
To construct the artificial dataset, we first obtained a binary representation for each node in the tree so that the Hamming distance between any pair of nodes is the same as the distance on the graph representation of the tree (Figure \ref{fig:binary_tree_schema}).
Let us call the set of binaries obtained this way by $A_0$.
We then generated a set of binaries, $A$,  by randomly flipping each coordinate value of $A_0$ with probability $\epsilon =0.1$.
The binary set $A$ was then embedded into $\mathbb{R}^d$ by mapping $a_1a_2...a_d$ to $[a_1, a_2, ..., a_d]$.
We used an Multi Layer Parceptron (MLP) of depth 3 and 100 hidden variables at each layer for both encoder and decoder of the VAE.
For activation function we used $tanh$.

\begin{table}[tbp]
  \centering
  \begin{tabular}{l|lcc}
    \toprule
    \multicolumn{2}{l}{Model} & Correlation & Correlation w/ noise \\
    \midrule
    \multirow{5}{*}{\rotatebox[origin=c]{90}{Vanilla}}
    & $\beta=0.1$ & $0.665 {\scriptstyle \pm .006}$ & $0.470 {\scriptstyle \pm .018}$ \\
    & $\beta=1.0$ & $0.644 {\scriptstyle \pm .007}$ & $0.550 {\scriptstyle \pm .012}$ \\
    & $\beta=2.0$ & $0.644 {\scriptstyle \pm .011}$ & $0.537 {\scriptstyle \pm .012}$ \\
    & $\beta=3.0$ & $0.638 {\scriptstyle \pm .004}$ & $0.501 {\scriptstyle \pm .044}$ \\
    & $\beta=4.0$ & $0.217 {\scriptstyle \pm .143}$ & $0.002 {\scriptstyle \pm .042}$ \\
    \multicolumn{2}{l}{Hyperbolic} & $\cellbest 0.768 {\scriptstyle \pm .003}$ & $\cellbest 0.590 {\scriptstyle \pm .018}$ \\
    \bottomrule
  \end{tabular}
  \caption{Results of tree embedding experiments for the Hyperbolic VAE and Vanilla VAEs trained with different weight constants for the KL term. We calculated the mean and the $\pm 1$ SD with five different experiments.}
  \label{tab:synthetic_binary_tree_result}
\end{table}

Table \ref{tab:synthetic_binary_tree_result} summarizes the quantitative comparison of Vanilla VAE against our Hyperbolic VAE.
For each pair of points in the tree, we computed their Hamming distance as well as their distance in the latent space of VAE.
That is, we used Hyperbolic distance for Hyperbolic VAE, and used Euclidean distance for Vanilla VAE.
We used the strength of correlation between the Hamming distances and the distances in the latent space as a measure of performance.
Hyperbolic VAE was performing better both on the original tree and on the artificial dataset generated from the tree.
Vanilla VAE performed the best with $\beta=2.0$, and collapsed with $\beta=3.0$.
The difference between Vanilla VAE and Hyperbolic VAE can be observed with much more clarity using the 2-dimensional visualization of the generated dataset on Poincar\'e Ball (See Figure \ref{fig:intro} and Appendix C.1).
The red points are the embeddings of $A_0$, and the blue points are the embeddings of all other points in $A$.
The pink $\times$ mark designates the origin of hyperbolic space.
For the visualization, we used the canonical diffeomorphism between the Lorenz model and the Poincar\'e ball model.

\subsection{MNIST}

\begin{table}[tbp]
  \centering
  \begin{tabular}{lcc}
    \toprule
    $n$ & Vanilla VAE & Hyperbolic VAE \\
    \midrule
    2  & $-140.45 {\scriptstyle \pm .47}$ & $\cellbest -138.61 {\scriptstyle \pm 0.45}$ \\
    5  & $-105.78 {\scriptstyle \pm .51}$ & $\cellbest -105.38 {\scriptstyle \pm 0.61}$ \\
    10  & $\cellbest -86.25 {\scriptstyle \pm .52}$ & $-86.40 {\scriptstyle \pm 0.28}$ \\
    20  & $\cellbest -77.89 {\scriptstyle \pm .36}$ & $-79.23 {\scriptstyle \pm 0.20}$ \\
    \bottomrule
  \end{tabular}
  \caption{Quantitative comparison of Hyperbolic VAE against Vanilla VAE on the MNIST dataset in terms of log-likelihood (LL) for several values of latent space dimension $n$. LL was computed using 500 samples of latent variables. We calculated the mean and the $\pm 1$ SD with five different experiments.
  }  \label{tab:mnist_result}
\end{table}

We applied Hyperbolic VAE to a binarized version of MNIST.
We used an MLP of depth 3 and 500 hidden units at each layer for both the encoder and the decoder of the VAE.
Table \ref{tab:mnist_result} shows the quantitative results of the experiments.
Log-likelihood was approximated with an empirical integration of the Bayesian predictor with respect to the latent variables \cite{Burda2016}.
Our method outperformed Vanilla VAE with small latent dimension.
Please see Appendix C.2 for further results.
\Cref{fig:mnist_sample} are the samples of the Hyperbolic VAE that was trained with 5-dimensional latent variables, and
\Cref{fig:mnist_embed} are the Poincar\'e Ball representations of the interpolations produced on $\mathbb{H}^2$ by the Hyperbolic VAE that was trained with 2-dimensional latent variables.

\begin{figure}[tbp]
  \vspace{-0.4cm}
  \centering
  \begin{tabular}{cc}

    \begin{minipage}[t]{0.45\hsize}
      \centering
      \subcaption{\leftline{}}
      \vspace{-0.2cm}
      \includegraphics[width=\hsize]{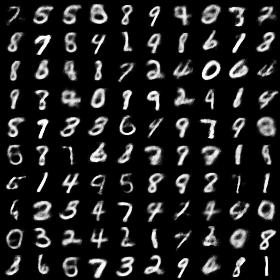}
      \label{fig:mnist_sample}
    \end{minipage}
    &
    \begin{minipage}[t]{0.45\hsize}
      \centering
      \subcaption{\leftline{}}
      \vspace{-0.2cm}
      \includegraphics[width=\hsize]{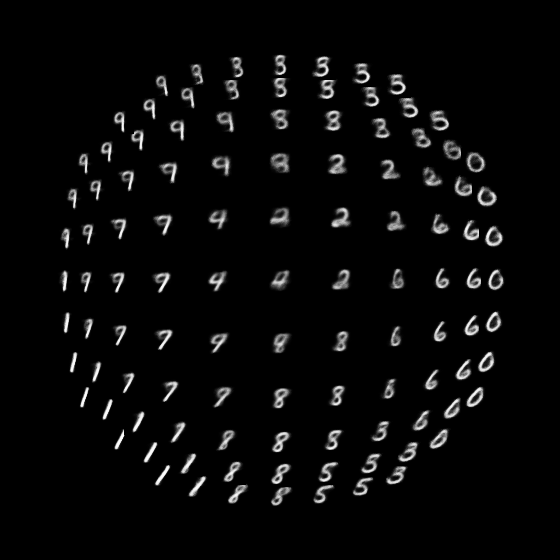}
      \label{fig:mnist_embed}
    \end{minipage}
  \end{tabular}
  \vspace{-0.4cm}
  \caption{(a) Samples generated from the Hyperbolic VAE trained on MNIST with latent dimension $n = 5$.
  (b): Interpolation of the MNIST dataset produced by the Hyperbolic VAE with latent dimension $n = 2$,  represented on the Poincar\'e ball.}
  \label{fig:vae_results}
\end{figure}

\subsection{Atari 2600 Breakout}
\begin{figure}[htbp]
  \centering
  \includegraphics[width=\hsize]{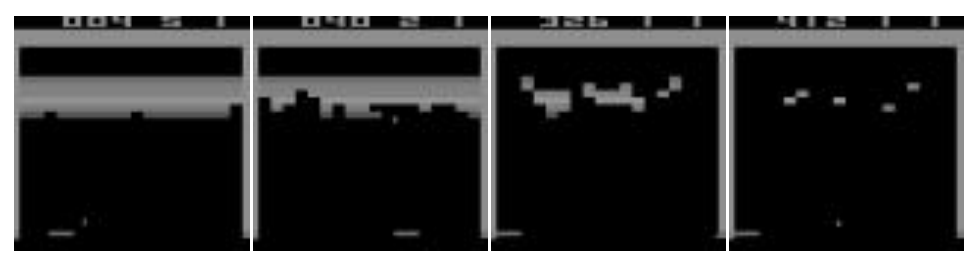}
  \caption{Examples of the observed screens in Atari 2600 Breakout.
  }
  \label{fig:breakout_dataset}
\end{figure}

In reinforcement learning, the number of possible state-action trajectories grows exponentially with the time horizon.
We may say that these trajectories often have a tree-like hierarchical structure that starts from the initial states.
We applied our Hyperbolic VAE to a set of trajectories that were explored by a trained policy during multiple episodes of Breakout in Atari 2600.
To collect the trajectories, we used a pretrained Deep Q-Network \cite{Mnih2015}, and used epsilon-greegy with $\epsilon= 0.1$.
We amassed a set of trajectories whose total length is 100,000, of which we used 80,000 as the training set, 10,000 as the validation set, and 10,000 as the test set.
Each frame in the dataset was gray-scaled and resized to 80 $\times$ 80.
The images in the Figure \ref{fig:breakout_dataset} are samples from the dataset.
We used a DCGAN-based architecture \cite{Radford2015} with latent space dimension $n=20$.
Please see Appendix D for more details.

\begin{figure*}[ht]
  \centering
  \includegraphics[width=\linewidth]{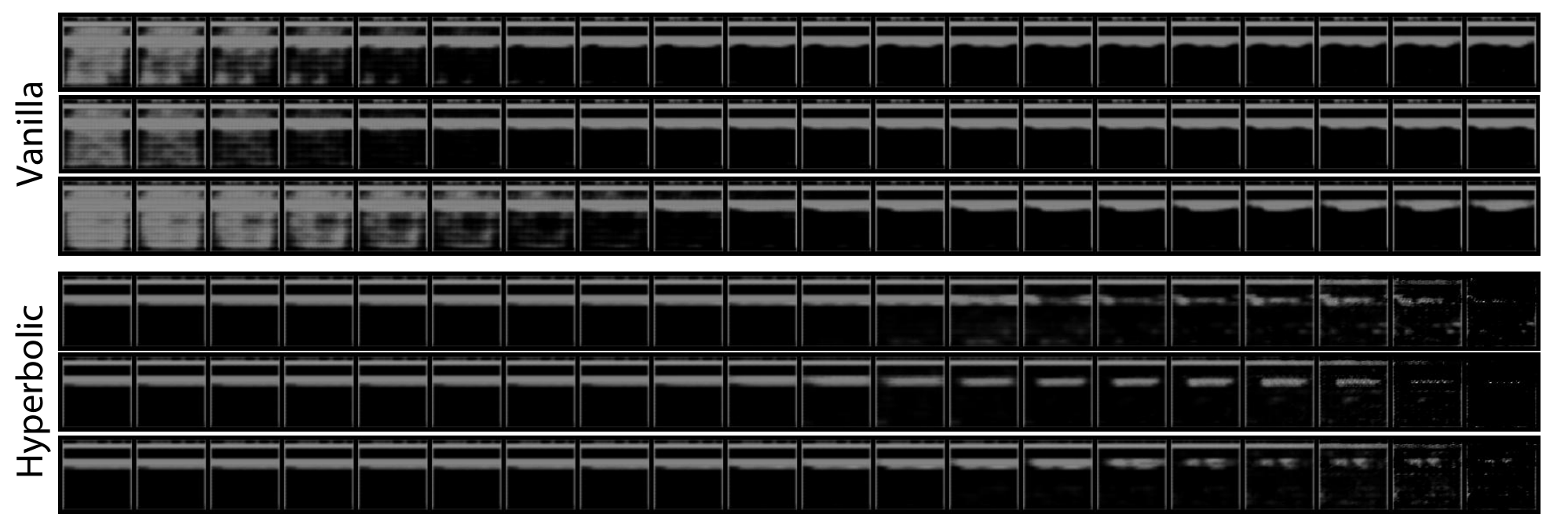}
  \caption{Samples from Vanilla and Hyperbolic VAEs trained on Atari 2600 Breakout screens. Each row was generated by sweeping the norm of $\tilde{\vb*{v}}$ from 1.0 to 10.0 in a log-scale.}
  \label{fig:breakout_norm_controlled_samples}
\end{figure*}

The Figure \ref{fig:breakout_norm_controlled_samples} is a visualization of our results.
The top three rows are the samples from Vanilla VAE, and the bottom three rows are the samples from Hyperbolic VAE.
Each row consists of samples generated from latent variables of the form $a \tilde{\vb*{v}} / \norm{\tilde{\vb*{v}}}_2$ with positive scalar $a$ in range $[1, 10]$.
Samples in each row are listed in increasing order of $a$.
For Vanilla VAE, we used $\mathcal{N}(\vb*{0}, I)$ as the prior.
For Hyperbolic VAE, we used $\mathcal{G}(\vb*{\mu}_0, I)$ as the prior.
We can see that the number of blocks decreases gradually and consistently in each row for Hyperbolic VAE.
Please see Appendix C.3 for more details and more visualizations.

In Breakout, the number of blocks is always finite, and blocks are located only in a specific region.
Let's refer to this specific region as $R$.
In order to evaluate each model-output based on the number of blocks,
we binarized each pixel in each output based on a prescribed luminance threshold and measured the proportion of the pixels with pixel value $1$ in the region $R$.
For each generated image,
we used this proportion as the measure of the number blocks contained in the image.

\begin{figure}[t]
  \centering
  \includegraphics[width=0.8\linewidth]{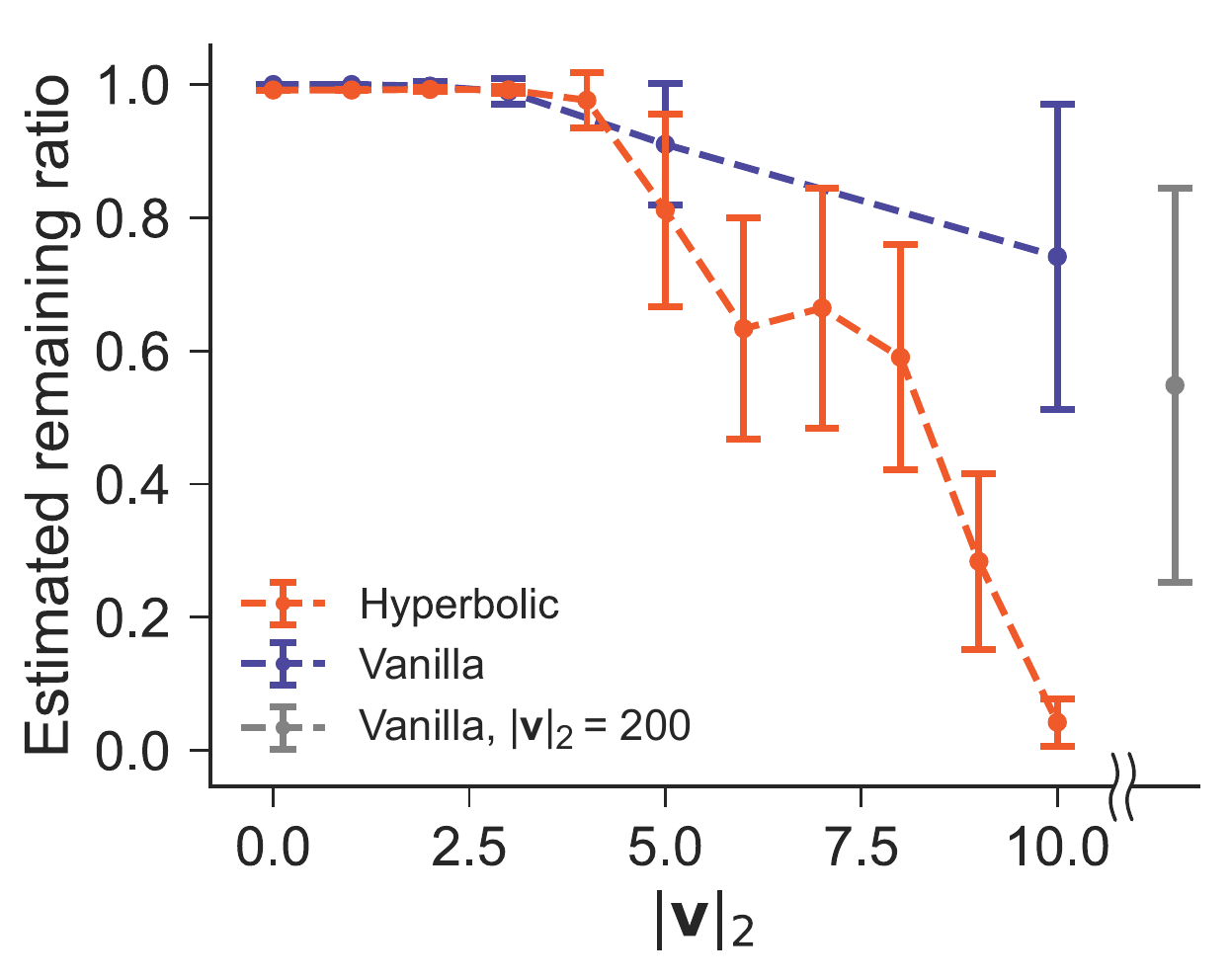}
  \vspace{-0.4cm}
  \caption{Estimated proportions of remaining blocks for Vanilla and Hyperbolic VAEs trained on Atari 2600 Breakout screens as they vary with the norm of latent variables sampled from a prior.}
  \label{fig:breakout_remaining_ratio}
\end{figure}

Figure \ref{fig:breakout_remaining_ratio} shows the estimated proportions of remaining blocks for Vanilla and Hyperbolic VAEs with different norm of $\tilde{\vb*{v}}$.
For Vanilla VAE, samples generated from $\tilde{\vb*{v}}$ with its norm as large as $\norm{\tilde{\vb*{v}}}_2=200$ contained considerable amount of blocks.
On the other hand, the number of blocks contained in a sample generated by Hyperbolic VAE decreased more consistently with the norm of $\norm{\tilde{\vb*{v}}}_2$.
This fact suggests that the cumulative reward up to a given state can be approximated well by the norm of Hyperbolic VAE's latent representation.
To validate this, we computed latent representation for each state in the test set and measured its correlation with the cumulative reward.
The correlation was 0.8460 for the Hyperbolic VAE.
For the Vanilla VAE, the correlation was 0.712.
We emphasize that no information regarding the reward was used during the training of both Vanilla and Hyperbolic VAEs.

\subsection{Word Embeddings}
\label{sub:word_embeddings_experiment}

Lastly, we applied pseudo-hyperbolic Gaussian to word embedding problem.
We trained probabilistic word embedding models with WordNet nouns dataset \cite{Miller1998} and evaluated the reconstruction performance of them (Table \ref{tab:wordnet_result}).
We followed the procedure of Poincar\'e embedding \cite{Nickel2017} and initialized all embeddings in the neighborhood of the origin.
In particular, we initialized each weight in the first linear part of the embedding by $\mathcal{N}(0, 0.01)$.
We treated the first 50 epochs as a burn-in phase and reduced the learning rate by a factor of $40$ after the burn-in phase.

In Table \ref{tab:wordnet_result}, 'Euclid' refers to the word embedding with Gaussian distribution on Euclidean space \cite{Vilnis2015}, and 'Hyperbolic' refers to our proposed method based on  pseudo-hyperbolic Gaussian.
Our hyperbolic model performed better than Vilnis' Euclidean counterpart when the latent space is low dimensional.
We used diagonal variance for both models above.
Please see Appendix C.4 for the full results.
We also performed the same experiment with unit variance.
The performance difference with small latent dimension was much more remarkable when we use unit variance.

\begingroup
\setlength{\tabcolsep}{4pt} 
\begin{table}[h]
  \centering
  \small
  \begin{tabular}{lcccc}
    \toprule
    & \multicolumn{2}{c}{Euclid} & \multicolumn{2}{c}{Hyperbolic} \\
    \cmidrule(l){2-5}
    $n$ & MAP & Rank & MAP & Rank \\
    \midrule
    5   & $0.296 {\scriptstyle \pm .006}$ & $25.09 {\scriptstyle \pm .80}$ & $\cellbest 0.506 {\scriptstyle \pm .017}$ & $\cellbest 20.55 {\scriptstyle \pm 1.34}$ \\
    10  & $0.778 {\scriptstyle \pm .007}$ & $\cellbest 4.70 {\scriptstyle \pm .05}$ & $\cellbest 0.795 {\scriptstyle \pm .007}$ & $5.07 {\scriptstyle \pm .12}$ \\
    20  & $0.894 {\scriptstyle \pm .002}$ & $\cellbest 2.23 {\scriptstyle \pm .03}$ & $\cellbest 0.897 {\scriptstyle \pm .005}$ & $2.54 {\scriptstyle \pm .20}$ \\
    50  & $0.942 {\scriptstyle \pm .003}$ & $1.51 {\scriptstyle \pm .04}$ & $\cellbest 0.975 {\scriptstyle \pm .001}$ & $\cellbest 1.19 {\scriptstyle \pm .01}$ \\
    100 & $0.953 {\scriptstyle \pm .002}$ & $1.34 {\scriptstyle \pm .02}$ & $\cellbest 0.978 {\scriptstyle \pm .002}$ & $\cellbest 1.15 {\scriptstyle \pm .01}$ \\
    \bottomrule
  \end{tabular}
  \caption{Experimental results of the reconstruction performance on the transitive closure of the WordNet noun hierarchy for several latent space dimension $n$. We calculated the mean and the $\pm 1$ SD with three different experiments.
  }\label{tab:wordnet_result}
\end{table}
\endgroup

\section{Conclusion}
\label{sec:conclusion}
In this paper, we proposed a novel parametrizaiton for the density of Gassusian on hyperbolic space that can both be differentiated and evaluated analytically.
Our experimental results on hyperbolic word embedding and hyperbolic VAE suggest that there is much more room left for the application of hyperbolic space.
Our parametrization enables gradient-based training of probabilistic models defined on hyperbolic space and opens the door to the investigation of complex models on hyperbolic space that could not have been explored before.

\section*{Acknowledgements}

We would like to thank Tomohiro Hayase, Kenta Oono, and Masaki Watanabe for helpful discussions.
We also thank Takeru Miyato and Sosuke Kobayashi for insightful reviews on the paper.
This paper is based on results obtained from Nagano's internship at Preferred Networks, Inc.
This work is partially supported by Masason Foundation.




\bibliography{hvae}

\begin{thebibliography}{25}
\providecommand{\natexlab}[1]{#1}
\providecommand{\url}[1]{\texttt{#1}}
\expandafter\ifx\csname urlstyle\endcsname\relax
  \providecommand{\doi}[1]{doi: #1}\else
  \providecommand{\doi}{doi: \begingroup \urlstyle{rm}\Url}\fi

\bibitem[Burda et~al.(2016)Burda, Grosse, and Salakhutdinov]{Burda2016}
Burda, Y., Grosse, R.~B., and Salakhutdinov, R.
\newblock Importance weighted autoencoders.
\newblock In \emph{Proceedings of the 4th International Conference on Learning
  Representations}, 2016.

\bibitem[Davidson et~al.(2018)Davidson, Falorsi, De~Cao, Kipf, and
  Tomczak]{Davidson2018}
Davidson, T.~R., Falorsi, L., De~Cao, N., Kipf, T., and Tomczak, J.~M.
\newblock Hyperspherical variational auto-encoders.
\newblock \emph{34th Conference on Uncertainty in Artificial Intelligence},
  2018.

\bibitem[Ganea et~al.(2018{\natexlab{a}})Ganea, B{\'{e}}cigneul, and
  Hofmann]{Ganea2018a}
Ganea, O., B{\'{e}}cigneul, G., and Hofmann, T.
\newblock Hyperbolic entailment cones for learning hierarchical embeddings.
\newblock In \emph{Proceedings of the 35th International Conference on Machine
  Learning}, pp.\  1632--1641, 2018{\natexlab{a}}.

\bibitem[Ganea et~al.(2018{\natexlab{b}})Ganea, B{\'{e}}cigneul, and
  Hofmann]{Ganea2018b}
Ganea, O., B{\'{e}}cigneul, G., and Hofmann, T.
\newblock In \emph{Advances in Neural Information Processing Systems 31}, pp.\
  5350--5360, 2018{\natexlab{b}}.

\bibitem[Grattarola et~al.(2018)Grattarola, Livi, and Alippi]{Grattarola2018}
Grattarola, D., Livi, L., and Alippi, C.
\newblock Adversarial autoencoders with constant-curvature latent manifolds.
\newblock \emph{CoRR}, abs/1812.04314, 2018.

\bibitem[Gretton et~al.(2012)Gretton, Borgwardt, Rasch, Sch\"{o}lkopf, and
  Smola]{Gretton2012}
Gretton, A., Borgwardt, K.~M., Rasch, M.~J., Sch\"{o}lkopf, B., and Smola, A.
\newblock A kernel two-sample test.
\newblock \emph{The Journal of Machine Learning Research}, 13:\penalty0
  723--773, March 2012.

\bibitem[G{\"{u}}l{\c{c}}ehre et~al.(2019)G{\"{u}}l{\c{c}}ehre, Denil,
  Malinowski, Razavi, Pascanu, Hermann, Battaglia, Bapst, Raposo, Santoro, and
  de~Freitas]{Gulcehre2018}
G{\"{u}}l{\c{c}}ehre, {\c{C}}., Denil, M., Malinowski, M., Razavi, A., Pascanu,
  R., Hermann, K.~M., Battaglia, P., Bapst, V., Raposo, D., Santoro, A., and
  de~Freitas, N.
\newblock Hyperbolic attention networks.
\newblock In \emph{Proceedings of the 7th International Conference on Learning
  Representations}, 2019.

\bibitem[Higgins et~al.(2017)Higgins, Matthey, Pal, Burgess, Glorot, Botvinick,
  Mohamed, and Lerchner]{Higgins2017}
Higgins, I., Matthey, L., Pal, A., Burgess, C., Glorot, X., Botvinick, M.,
  Mohamed, S., and Lerchner, A.
\newblock $\beta$-vae: Learning basic visual concepts with a constrained
  variational framework.
\newblock In \emph{Proceedings of the 5th International Conference on Learning
  Representations}, 2017.

\bibitem[Jang et~al.(2017)Jang, Gu, and Poole]{Jang2017}
Jang, E., Gu, S., and Poole, B.
\newblock Categorical reparameterization with gumbel-softmax.
\newblock In \emph{Proceedings of the 5th International Conference on Learning
  Representations}, 2017.

\bibitem[Kingma \& Welling(2014)Kingma and Welling]{Kingma2013}
Kingma, D.~P. and Welling, M.
\newblock Auto-encoding variational bayes.
\newblock In \emph{Proceedings of the 2nd International Conference on Learning
  Representations}, 2014.

\bibitem[Kingma et~al.(2016)Kingma, Salimans, Jozefowicz, Chen, Sutskever, and
  Welling]{Kingma2016}
Kingma, D.~P., Salimans, T., Jozefowicz, R., Chen, X., Sutskever, I., and
  Welling, M.
\newblock Improved variational inference with inverse autoregressive flow.
\newblock In \emph{Advances in Neural Information Processing Systems 29}, pp.\
  4743--4751. 2016.

\bibitem[Mikolov et~al.(2013)Mikolov, Sutskever, Chen, Corrado, and
  Dean]{Mikolov2013}
Mikolov, T., Sutskever, I., Chen, K., Corrado, G.~S., and Dean, J.
\newblock Distributed representations of words and phrases and their
  compositionality.
\newblock In \emph{Advances in Neural Information Processing Systems 26}, pp.\
  3111--3119. 2013.

\bibitem[Miller(1998)]{Miller1998}
Miller, G.
\newblock \emph{WordNet: An electronic lexical database}.
\newblock MIT press, 1998.

\bibitem[Mnih et~al.(2015)Mnih, Kavukcuoglu, Silver, Rusu, Veness, Bellemare,
  Graves, Riedmiller, Fidjeland, Ostrovski, et~al.]{Mnih2015}
Mnih, V., Kavukcuoglu, K., Silver, D., Rusu, A.~A., Veness, J., Bellemare,
  M.~G., Graves, A., Riedmiller, M., Fidjeland, A.~K., Ostrovski, G., et~al.
\newblock Human-level control through deep reinforcement learning.
\newblock \emph{Nature}, 518\penalty0 (7540):\penalty0 529, 2015.

\bibitem[Nickel \& Kiela(2017)Nickel and Kiela]{Nickel2017}
Nickel, M. and Kiela, D.
\newblock Poincar\'{e} embeddings for learning hierarchical representations.
\newblock In \emph{Advances in Neural Information Processing Systems 30}, pp.\
  6338--6347. 2017.

\bibitem[Nickel \& Kiela(2018)Nickel and Kiela]{Nickel2018}
Nickel, M. and Kiela, D.
\newblock Learning continuous hierarchies in the lorentz model of hyperbolic
  geometry.
\newblock In \emph{Proceedings of the 35th International Conference on Machine
  Learning}, pp.\  3776--3785, 2018.

\bibitem[Ovinnikov(2019)]{Ovinnikov2019}
Ovinnikov, I.
\newblock Poincar\'e wasserstein autoencoder.
\newblock \emph{CoRR}, abs/1901.01427, 2019.

\bibitem[Radford et~al.(2016)Radford, Metz, and Chintala]{Radford2015}
Radford, A., Metz, L., and Chintala, S.
\newblock Unsupervised representation learning with deep convolutional
  generative adversarial networks.
\newblock In \emph{Proceedings of the 4th International Conference on Learning
  Representations}, 2016.

\bibitem[Rezende \& Mohamed(2015)Rezende and Mohamed]{Rezende2015}
Rezende, D.~J. and Mohamed, S.
\newblock Variational inference with normalizing flows.
\newblock In \emph{Proceedings of the 32nd International Conference on Machine
  Learning}, pp.\  1530--1538, 2015.

\bibitem[Rezende et~al.(2014)Rezende, Mohamed, and Wierstra]{Rezende2014}
Rezende, D.~J., Mohamed, S., and Wierstra, D.
\newblock Stochastic backpropagation and approximate inference in deep
  generative models.
\newblock In \emph{Proceedings of the 31st International Conference on Machine
  Learning}, volume~32, pp.\  1278--1286, 2014.

\bibitem[Rolfe(2017)]{Rolfe2017}
Rolfe, J.~T.
\newblock Discrete variational autoencoders.
\newblock In \emph{Proceedings of the 5th International Conference on Learning
  Representations}, 2017.

\bibitem[Sala et~al.(2018)Sala, De~Sa, Gu, and Re]{Sala2018}
Sala, F., De~Sa, C., Gu, A., and Re, C.
\newblock Representation tradeoffs for hyperbolic embeddings.
\newblock In \emph{Proceedings of the 35th International Conference on Machine
  Learning}, volume~80, pp.\  4460--4469, 2018.

\bibitem[Sarkar(2012)]{Sarkar2012}
Sarkar, R.
\newblock Low distortion delaunay embedding of trees in hyperbolic plane.
\newblock In \emph{Graph Drawing}, pp.\  355--366, 2012.

\bibitem[Vikram et~al.(2018)Vikram, Hoffman, and Johnson]{Vikram2018}
Vikram, S., Hoffman, M.~D., and Johnson, M.~J.
\newblock The loracs prior for vaes: Letting the trees speak for the data.
\newblock \emph{CoRR}, abs/1810.06891, 2018.

\bibitem[Vilnis \& McCallum(2015)Vilnis and McCallum]{Vilnis2015}
Vilnis, L. and McCallum, A.
\newblock Word representations via gaussian embedding.
\newblock In \emph{Proceedings of the 3rd International Conference on Learning
  Representations}, 2015.

\end{thebibliography}
\bibliographystyle{styles/icml2019}





\clearpage

\onecolumn

\icmltitle{Appendix: A Wrapped Normal Distribution on Hyperbolic Space for Gradient-Based Learning}

\icmlkeywords{Machine Learning, ICML}

\vskip 0.3in

\renewcommand{\thesection}{\Alph{section}}
\setcounter{section}{0}

\section{Derivations}
\subsection{Inverse Exponential Map}
As we mentioned in the main text, the exponential map from $T_{\vb*{\mu}} \mathbb{H}^n$ to $\mathbb{H}^n$ is given by
\begin{equation}
  \vb*{z} = \exp_{\vb*{\mu}}(\vb*{u}) = \cosh \qty( \norm{\vb*{u}}_{\mathcal{L}} ) \vb*{\mu} + \sinh \qty( \norm{\vb*{u}}_{\mathcal{L}} ) \frac{\vb*{u}}{\norm{\vb*{u}}_{\mathcal{L}}}. \label{eq:expmap}
\end{equation}

\noindent
Solving \eqref{eq:expmap} for $\vb*{u}$, we obtain
\begin{equation*}
  \vb*{u} = \frac{\norm{\vb*{u}}_{\mathcal{L}}}{\sinh(\norm{\vb*{u}}_{\mathcal{L}})} \qty(\vb*{z} - \cosh (\norm{\vb*{u}}_{\mathcal{L}}) \vb*{\mu}).
\end{equation*}
We still need to obtain the evaluatable expression for $\norm{\vb*{u}}_{\mathcal{L}}$.
Using the characterization of the tangent space (main text, (2)),  we see that
\begin{align*}
  \langle \vb*{\mu}, \vb*{u} \rangle_{\mathcal{L}} =& \frac{\norm{\vb*{u}}_{\mathcal{L}}}{\sinh(\norm{\vb*{u}}_{\mathcal{L}})} \qty\Big( \langle \vb*{\mu}, \vb*{z} \rangle_{\mathcal{L}} - \cosh (\norm{\vb*{u}}_{\mathcal{L}}) \langle \vb*{\mu}, \vb*{\mu} \rangle_{\mathcal{L}} ) = 0, \\
  \cosh(\norm{\vb*{u}}_{\mathcal{L}}) =& - \langle \vb*{\mu}, \vb*{z} \rangle_{\mathcal{L}}, \\
  \norm{\vb*{u}}_{\mathcal{L}} =& \arccosh (- \langle \vb*{\mu}, \vb*{z} \rangle_{\mathcal{L}}).
\end{align*}
Now, defining
\noindent
$\alpha = - \langle \vb*{\mu}, \vb*{z} \rangle_{\mathcal{L}}$,
we can obtain the inverse exponential function as
\begin{equation}
  \vb*{u} = \exp^{-1}_{\vb*{\mu}}(\vb*{z}) = \frac{ \arccosh(\alpha) }{ \sqrt{\alpha^2 - 1} } \qty(\vb*{z} - \alpha \vb*{\mu}). \label{eq:inv_expmap}
\end{equation}.

\subsection{Inverse Parallel Transport}
The parallel transportation on the Lorentz model along the geodesic from $\vb*{\nu}$ to $\vb*{\mu}$ is given by
\begin{align}
  \PT_{\vb*{\nu} \rightarrow \vb*{\mu}} (\vb*{v}) &= \vb*{v} - \frac{ \langle \exp^{-1}_{\vb*{\nu}}(\vb*{\mu}), \vb*{v} \rangle_{\mathcal{L}} }{ d_{\ell} (\vb*{\nu}, \vb*{\mu})^2 } \qty( \exp^{-1}_{\vb*{\nu}}(\vb*{\mu}) + \exp^{-1}_{\vb*{\mu}}(\vb*{\nu}) ) \nonumber \\
  &= \vb*{v} + \frac{ \langle \vb*{\mu} - \alpha \vb*{\nu}, \vb*{v} \rangle_{\mathcal{L}} }{\alpha + 1} \qty( \vb*{\nu} + \vb*{\mu} ), \label{eq:parallel}
\end{align}
where $\alpha = - \langle \vb*{\nu}, \vb*{\mu} \rangle_{\mathcal{L}}$.
Next, likewise, for the exponential map, we need to be able to compute the inverse of the parallel transform.
Solving \eqref{eq:parallel} for $\vb*{v}$, we get
\vspace{-0.1cm}

\begin{equation*}
  \vb*{v} = \vb*{u} - \frac{ \langle \vb*{\mu} - \alpha \vb*{\nu}, \vb*{v} \rangle_{\mathcal{L}} }{\alpha + 1} (\vb*{\nu} + \vb*{\mu}).
\end{equation*}

\noindent
Now, observing that
\vspace{-0.1cm}

\begin{align*}
  \langle \vb*{\nu} - \alpha \vb*{\mu}, \vb*{u} \rangle_{\mathcal{L}} =& \langle \vb*{\nu}, \vb*{v} \rangle_{\mathcal{L}} + \frac{ \langle \vb*{\mu} - \alpha \vb*{\nu}, \vb*{v} \rangle_{\mathcal{L}} }{ \alpha + 1 } \qty( \langle \vb*{\nu}, \vb*{\nu} \rangle_{\mathcal{L}} + \langle \vb*{\mu}, \vb*{\nu} \rangle_{\mathcal{L}} ) \\
  =& - \langle \vb*{\mu}, \vb*{v} \rangle_{\mathcal{L}} = - \langle \vb*{\mu} - \alpha \vb*{\nu}, \vb*{v} \rangle_{\mathcal{L}},
\end{align*}
\noindent
we can write the inverse parallel transport as
\vspace{-0.1cm}

\begin{equation}
  \vb*{v} = \PT_{\vb*{\nu} \rightarrow \vb*{\mu}}^{-1} (\vb*{u}) = \vb*{u} + \frac{ \langle \vb*{\nu} - \alpha \vb*{\mu}, \vb*{u} \rangle_{\mathcal{L}} }{\alpha + 1} (\vb*{\nu} + \vb*{\mu}) \label{eq:inv_parallel}.
\end{equation}

\noindent
The inverse of parallel transport from $\vb*{\nu}$ to $\vb*{\mu}$ coincides with the parallel transport from $\vb*{\mu}$ to $\vb*{\nu}$.

\subsection{Determinant of exponential map}

We provide the details of the computation of the log determinant of $\exp_\mu$.
Let $\vb*{\mu} \in \mathbb{H}^n$, let $\vb*{u} \in T_\mu (\mathbb{H}^{n})$, and let $\vb*{v} = \exp_\mu(\vb*{u})$.
Then the derivative is a map from the tangent space of  $T_\mu (\mathbb{H}^n)$ at $\vb*{u}$ to the tangent space of $\mathbb{H}^n$ at $\vb*{v}$.
The determinant of this derivative will not change by any orthogonal change of basis.
Let us choose an orthonormal basis of $T_u (T_\mu (\mathbb{H}^{n})) \cong T_\mu (\mathbb{H}^{n})$ containing $\bar{\vb*{u}} = \vb*{u}/\norm{\vb*{u}}_{\mathcal{L}}$:
$$\{\bar{\vb*{u}}, \vb*{u}^\prime_1, \vb*{u}^\prime_2, ... ,\vb*{u}^\prime_{n-1} \} $$
The desired determinant can be computed by tracking how much each element of this basis grows in magnitude under the transformation.

The derivative in the direction of each basis element can be computed as follows:
\begin{align}
  \dd \exp_{\vb*{\mu}}(\bar{\vb*{u}}) =& \pdv{\epsilon} \bigg|_{\epsilon = 0} \nonumber \exp_{\vb*{\mu}} (\vb*{u} + \epsilon \bar{\vb*{u}}) \nonumber \\
  =& \pdv{\epsilon} \bigg|_{\epsilon = 0} \qty[ \cosh(r + \epsilon) \vb*{\mu} + \sinh(r + \epsilon) \frac{ \vb*{u} + \epsilon \bar{\vb*{u}} }{ \norm{ \vb*{u} + \epsilon \bar{\vb*{u}} }_{\mathcal{L}} } ] \nonumber \\
  =& \sinh(r) \vb*{\mu} + \cosh(r) \bar{\vb*{u}}. \label{eq:dexp_para}
\end{align}

\begin{align}
  \dd \exp_{\vb*{\mu}}(\vb*{u}^\prime_k) =& \pdv{\epsilon} \bigg|_{\epsilon = 0} \nonumber \exp_{\vb*{\mu}} (\vb*{u} + \epsilon \vb*{u}^\prime_k) \nonumber \\
  =& \pdv{\epsilon} \bigg|_{\epsilon = 0} \qty[ \cosh(r) \vb*{\mu} + \sinh(r) \frac{ \vb*{u} + \epsilon \vb*{u}^\prime }{ r } ] \nonumber \\
  =& \frac{\sinh r}{r} \vb*{u}^\prime. \label{eq:dexp_perp}
\end{align}
In the second line of the computation of the directional derivative with respect to $\vb*{u^\prime}_k$, we used the fact that
\noindent
$\norm{\vb*{u} + \epsilon \vb*{u}^\prime_k}_{\mathcal{L}} = \sqrt{ \langle \vb*{u}, \vb*{u} \rangle_{\mathcal{L}} + \epsilon \langle \vb*{u}, \vb*{u}^\prime_k \rangle_{\mathcal{L}} + \epsilon^2 \langle \vb*{u}^\prime_k, \vb*{u}^\prime_k \rangle_{\mathcal{L}} } = \norm{\vb*{u}}_{\mathcal{L}} + \mathcal{O}(\epsilon^2)$
and that $O(\epsilon^2)$ in the above expression will disappear in the $\epsilon \to 0 $ limit of the finite difference.
All together, the derivatives computed with respect to our choice of the basis elements are given by
\begin{equation}
  \qty(\sinh(r) \vb*{\mu} + \cosh(r) \bar{\vb*{u}}, \frac{\sinh r}{r} \vb*{u}^\prime_1, \: \frac{\sinh r}{r} \vb*{u}^\prime_2, \: \cdots, \: \frac{\sinh r}{r} \vb*{u}^\prime_{n-1}) \nonumber
\end{equation}

\noindent
The desired determinant is the product of the Lorentzian norms of the vectors of the set above.
Because all elements of $T_\mu(\mathcal{H}^{n}) \subset \mathbb{R}^{n}$ are orthogonal with respect to the Lorentzian inner product and because  $\norm{\sinh(r) \vb*{\mu} + \cosh(r) \bar{\vb*{u}}}_{\mathcal{L}}=1$ and $\norm{\sinh(r) / r \cdot \vb*{u}^\prime}_{\mathcal{L}}=\sinh(r) / r$, we get
\begin{equation}
  \det \qty( \pdv{\exp_{\vb*{\mu}}(\vb*{u})}{\vb*{u}}) = \qty(\frac{\sinh r}{r})^{n-1}.
\end{equation}

\subsection{Determinant of parallel transport}
Next, let us compute the determinant of the parallel transport.
Let $\vb*{v} \in T_{\vb*{\mu}_0} \mathbb{H}^n$, and let
$\vb*{u}=\PT_{\vb*{\mu}_0 \rightarrow \vb*{\mu}}(\vb*{v}) \in T_{\vb*{\mu}} \mathbb{H}^n$.
The derivative of this map is a map from $T_v ( T_{\mu_0} (\mathbb{H}^n))$
to $T_u(\mathbb{H}^n)$.
Let us choose an orthonormal basis $\vb*{\xi}_k$ (In Lorentzian sense).
Likewise above, we can compute the desired determinant by tracking how much each element of this basis grows in magnitude under the transformation.

Denoting $\alpha = - \langle \vb*{\mu}_0, \vb*{\mu} \rangle_{\mathcal{L}}$,
we get
\begin{align}
  \dd \PT_{\vb*{\mu}_0 \rightarrow \vb*{\mu}}(\vb*{\xi}) =& \pdv{\epsilon} \bigg|_{\epsilon = 0} \nonumber \PT_{\vb*{\mu}_0 \rightarrow \vb*{\mu}} (\vb*{v} + \epsilon \vb*{\xi}) \nonumber \\
  =& \pdv{\epsilon} \bigg|_{\epsilon = 0} \qty[ (\vb*{v} + \epsilon \vb*{\xi}) + \frac{ \langle \vb*{\mu} - \alpha \vb*{\mu}_0, \vb*{v} + \epsilon \vb*{\xi} \rangle_{\mathcal{L}} }{\alpha + 1} (\vb*{\mu}_0 + \vb*{\mu}) ] \nonumber \\
  =& \vb*{\xi} + \frac{ \langle \vb*{\mu} - \alpha \vb*{\mu}_0, \vb*{\xi} \rangle_{\mathcal{L}} }{ \alpha + 1 } (\vb*{\mu}_0 + \vb*{\mu}) = \PT_{\vb*{\mu}_0 \rightarrow \vb*{\mu}}(\vb*{\xi}). \label{eq:dPT}
\end{align}
and see that each basis element $\vb*{\xi}_k$ is mapped by $\dd \PT_{\vb*{\mu}_0 \rightarrow \vb*{\mu}}$ to
\noindent

\begin{equation}
  \qty(\PT_{\vb*{\mu}_0 \rightarrow \vb*{\mu}}(\vb*{\xi}_1), \PT_{\vb*{\mu}_0 \rightarrow \vb*{\mu}}(\vb*{\xi}_2) \: \cdots, \: \PT_{\vb*{\mu}_0 \rightarrow \vb*{\mu}}(\vb*{\xi}_n)) \nonumber
\end{equation}
\noindent
Because parallel transport is a norm preserving map, $\norm{ \PT_{\vb*{\mu}_0 \rightarrow \vb*{\mu}}(\vb*{\xi}) }_{\mathcal{L}} = 1$. That is,
\begin{equation}
  \det \qty( \pdv{ \PT_{ \vb*{\mu}_0 \rightarrow \vb*{\mu}}(\vb*{v}) }{ \vb*{v} } ) = 1.
\end{equation}

\newpage

\section{Visual Examples of Pseudo-Hyperbolic Gaussian $\mathcal{G}(\vb*{\mu}, \Sigma)$}

\begin{figure}[htbp]
  \centering
  \begin{tabular}{cc}

    \begin{minipage}[t]{0.32\hsize}
      \centering
      \subcaption{\leftline{}}
      \vspace{-0.3cm}
      \includegraphics[width=\linewidth]{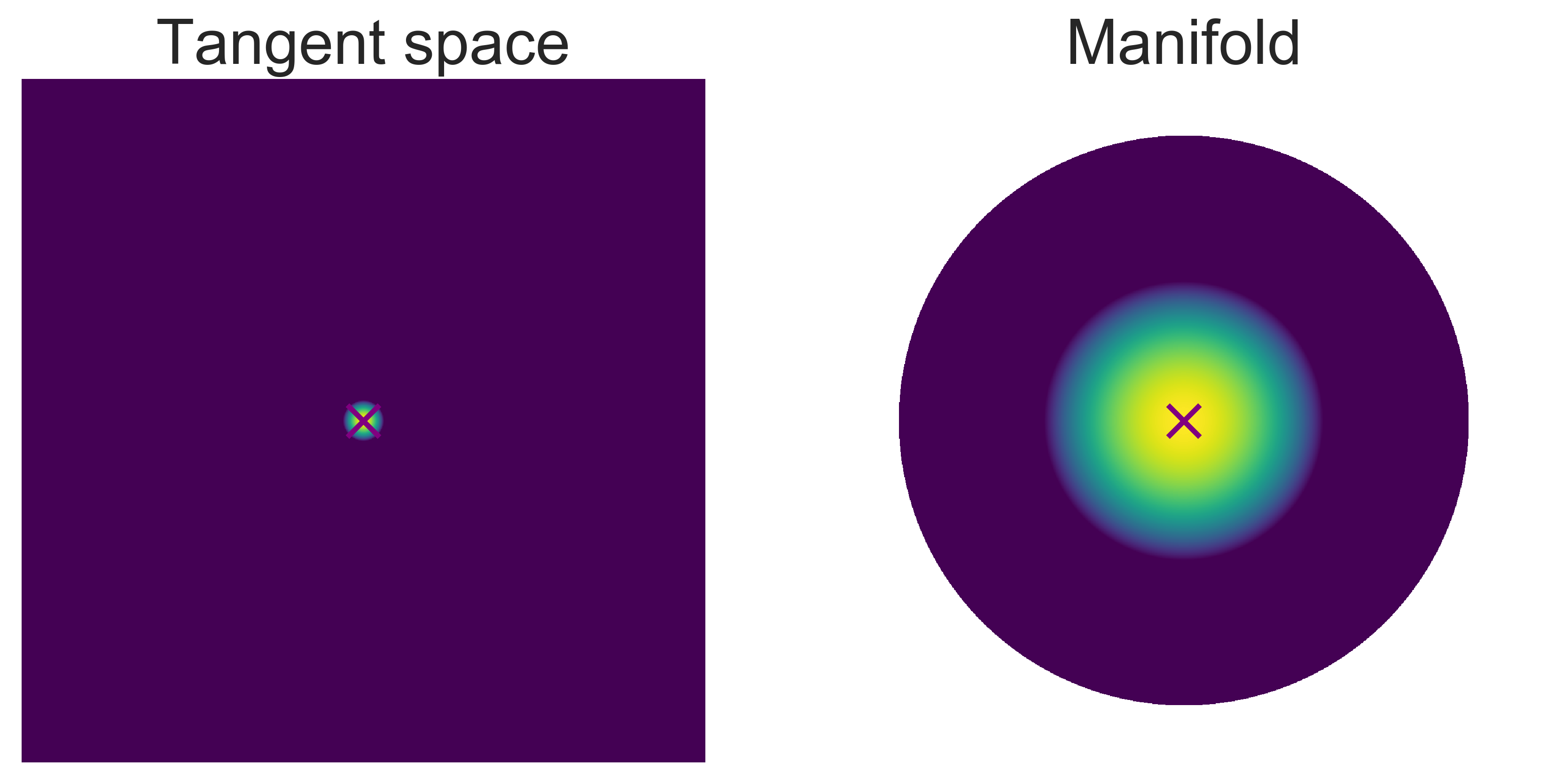}
    \end{minipage}
    &
    \begin{minipage}[t]{0.32\hsize}
      \centering
      \subcaption{\leftline{}}
      \vspace{-0.3cm}
      \includegraphics[width=\linewidth]{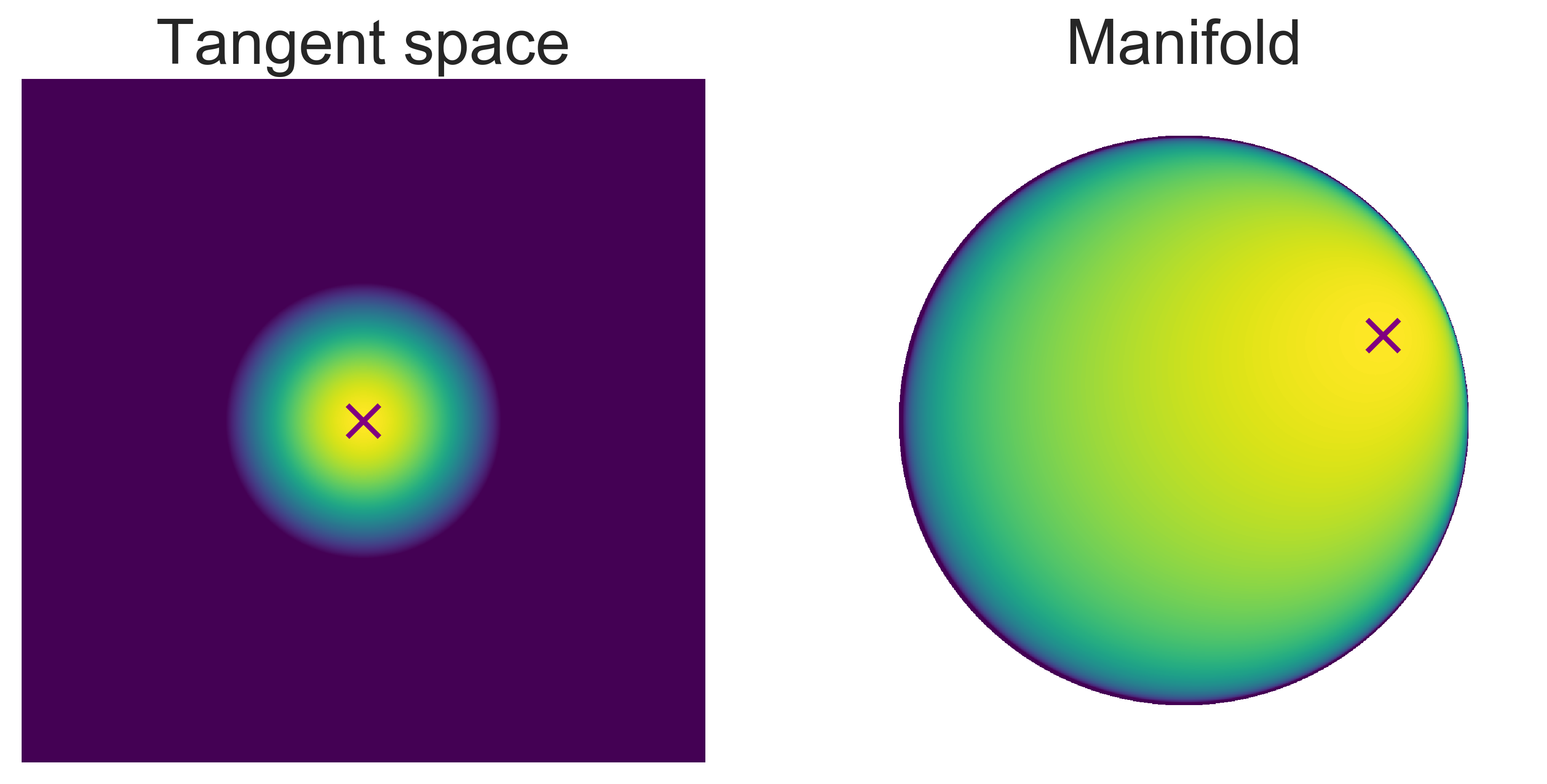}
    \end{minipage}
    \\
    \begin{minipage}[t]{0.32\hsize}
      \centering
      \subcaption{\leftline{}}
      \vspace{-0.3cm}
      \includegraphics[width=\linewidth]{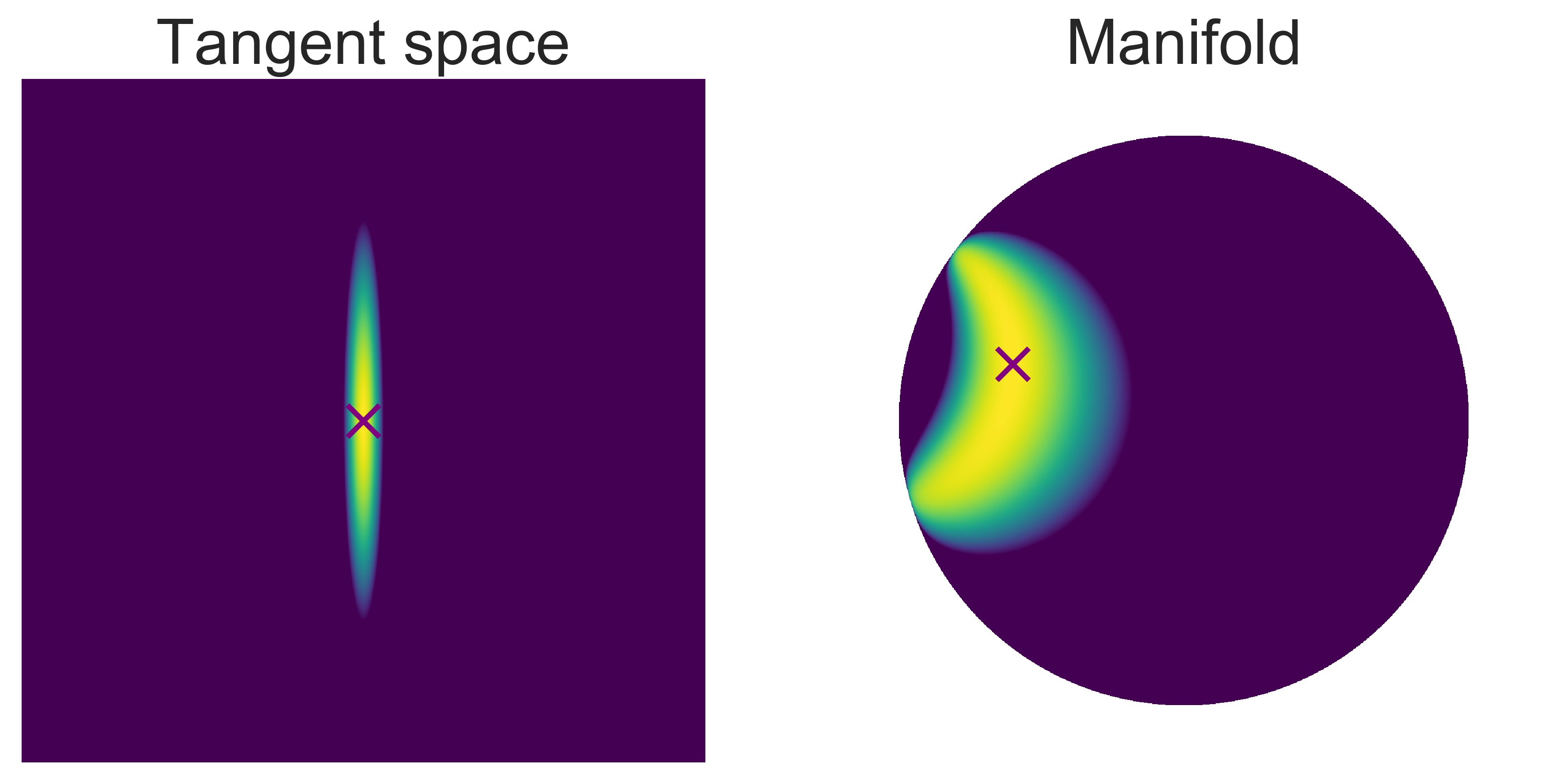}
    \end{minipage}
    &
    \begin{minipage}[t]{0.32\hsize}
      \centering
      \subcaption{\leftline{}}
      \vspace{-0.3cm}
      \includegraphics[width=\linewidth]{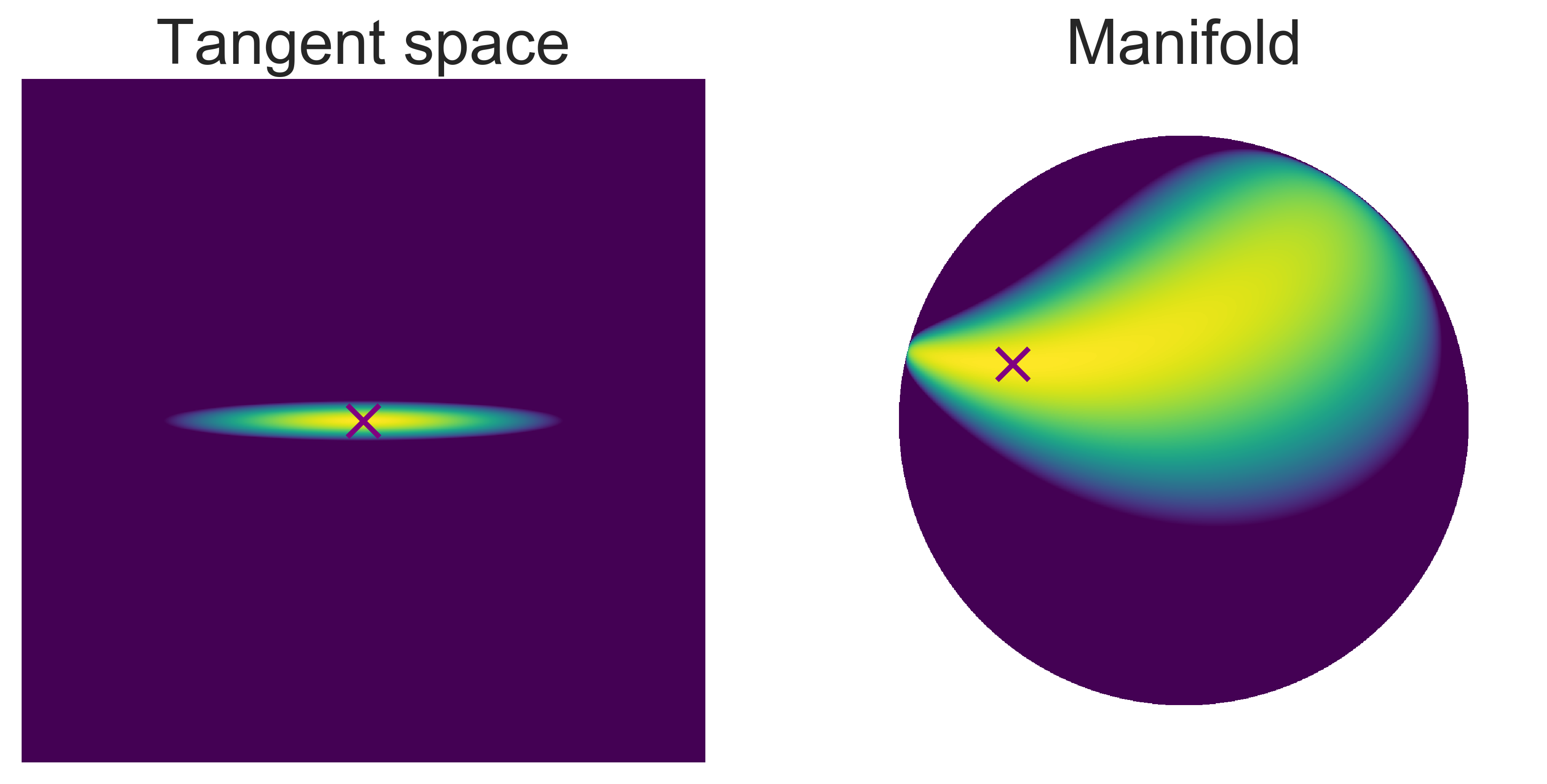}
    \end{minipage}
    \\
    \begin{minipage}[t]{0.32\hsize}
      \centering
      \subcaption{\leftline{}}
      \vspace{-0.3cm}
      \includegraphics[width=\linewidth]{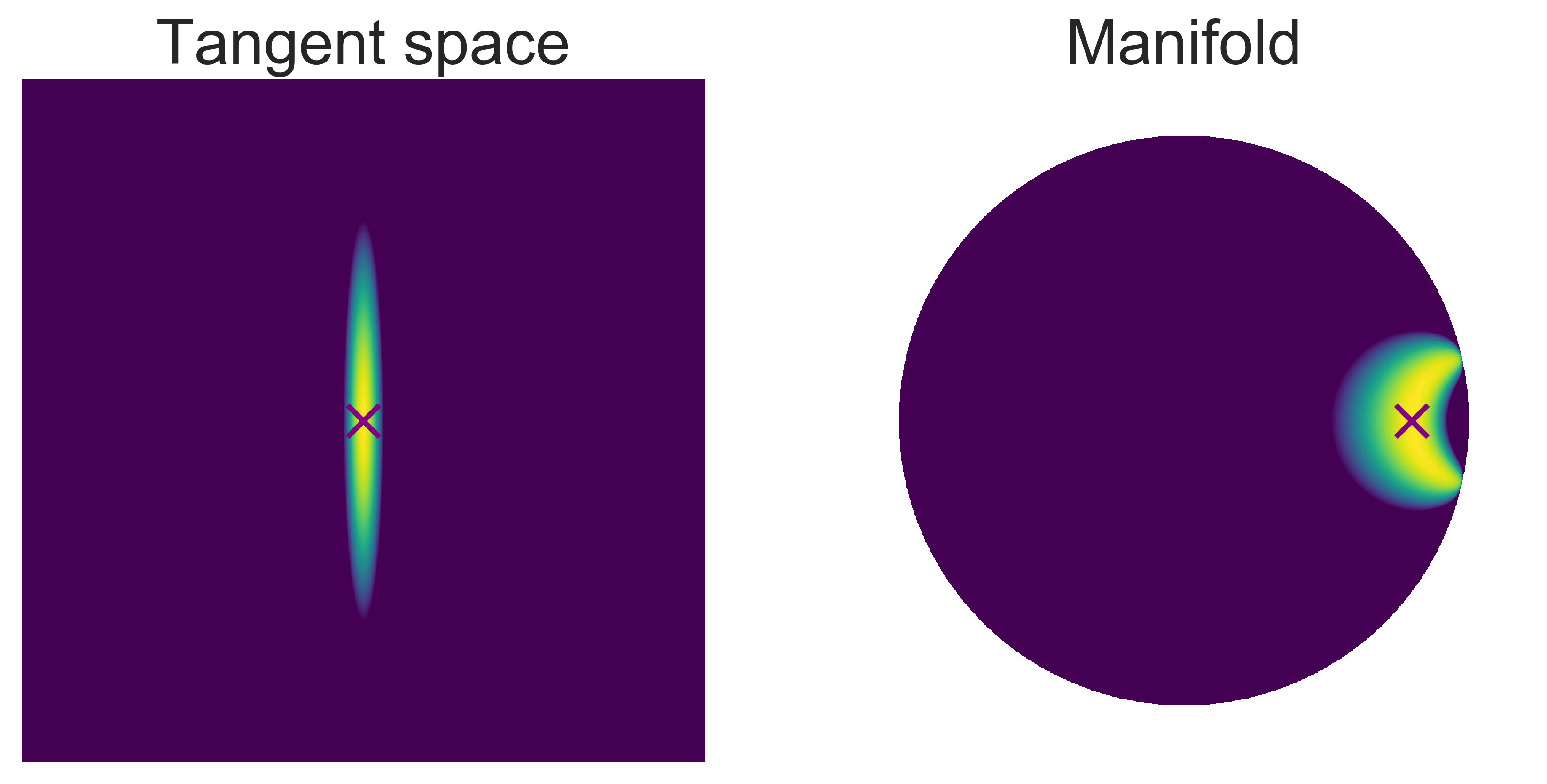}
    \end{minipage}
    &
    \begin{minipage}[t]{0.32\hsize}
      \centering
      \subcaption{\leftline{}}
      \vspace{-0.3cm}
      \includegraphics[width=\linewidth]{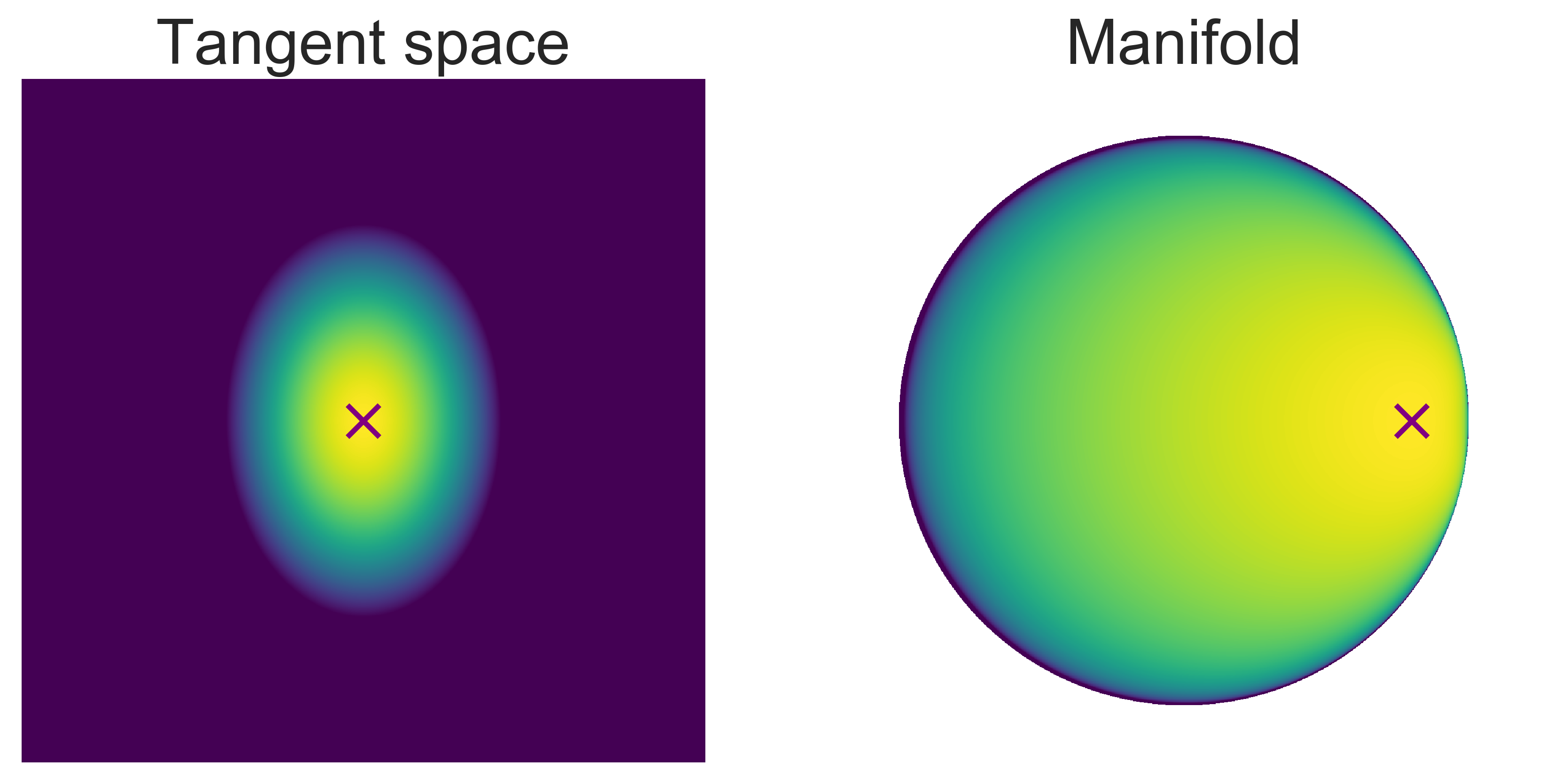}
    \end{minipage}
    \\
    \begin{minipage}[t]{0.32\hsize}
      \centering
      \subcaption{\leftline{}}
      \vspace{-0.3cm}
      \includegraphics[width=\linewidth]{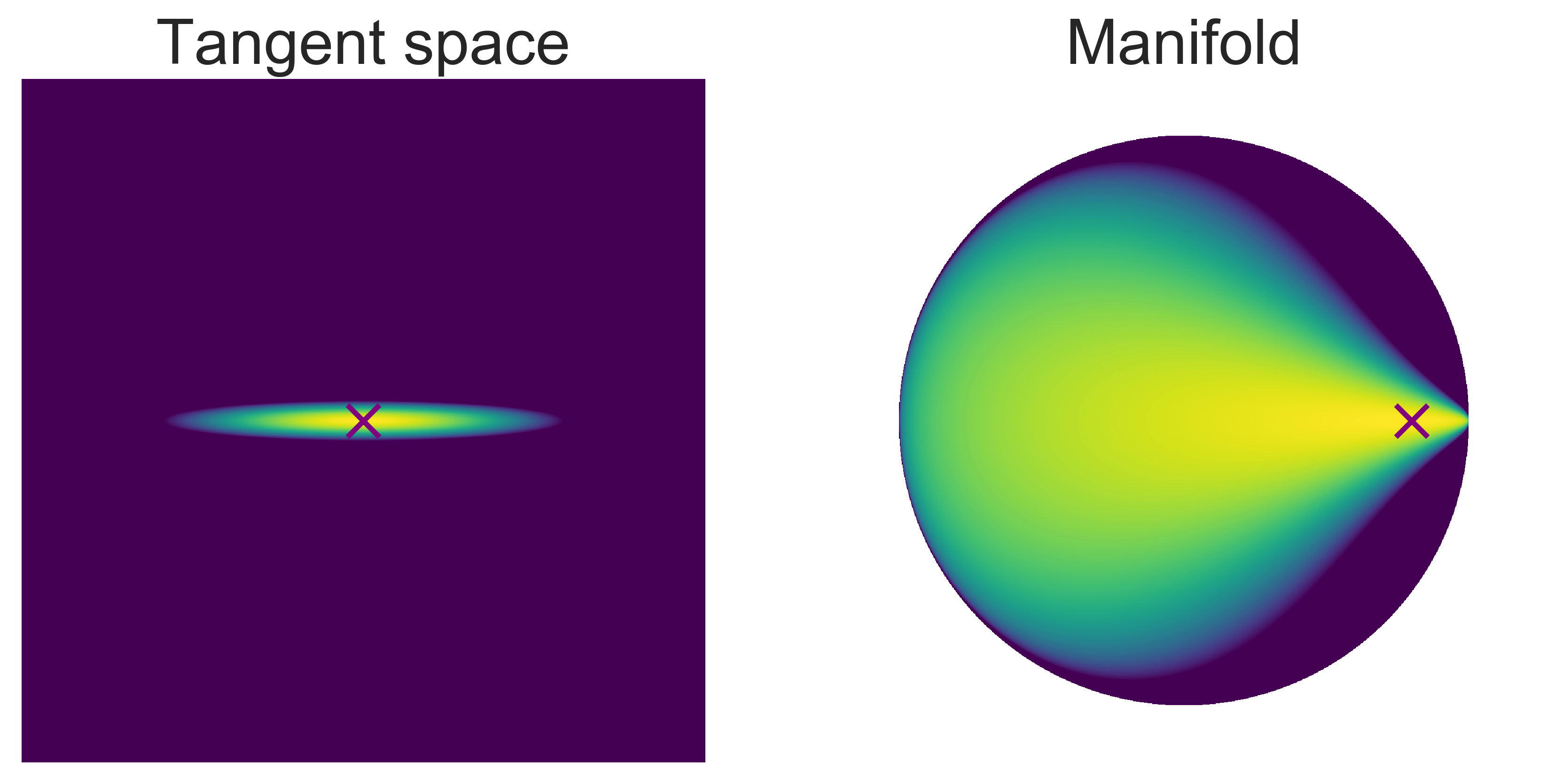}
    \end{minipage}
    &
    \begin{minipage}[t]{0.32\hsize}
      \centering
      \subcaption{\leftline{}}
      \vspace{-0.3cm}
      \includegraphics[width=\linewidth]{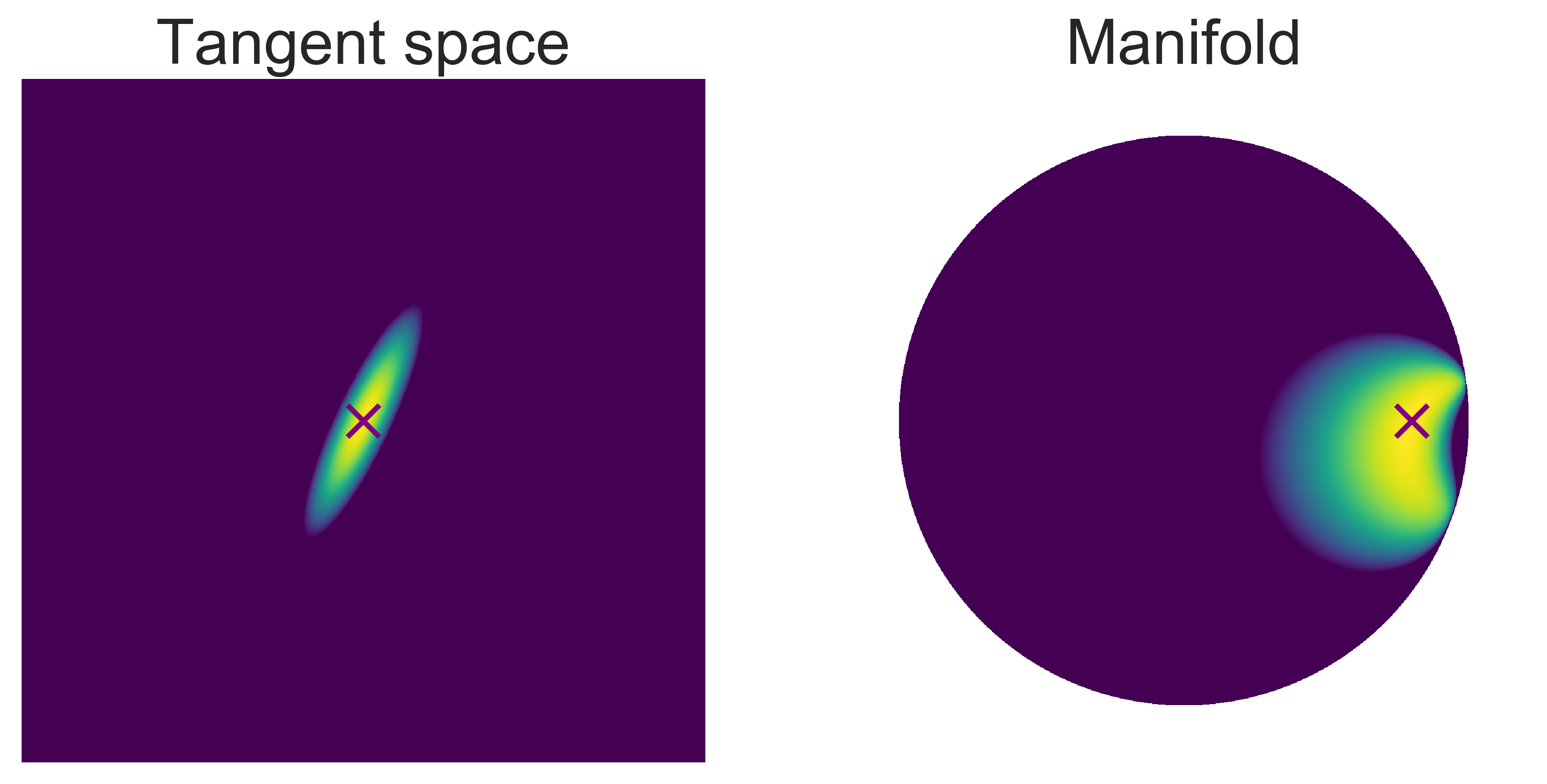}
    \end{minipage}
    \\
    \begin{minipage}[t]{0.32\hsize}
      \centering
      \subcaption{\leftline{}}
      \vspace{-0.3cm}
      \includegraphics[width=\linewidth]{{images/densities/density_withcov_summary_09}.png}
    \end{minipage}
    &
    \begin{minipage}[t]{0.32\hsize}
      \centering
      \subcaption{\leftline{}}
      \vspace{-0.3cm}
      \includegraphics[width=\linewidth]{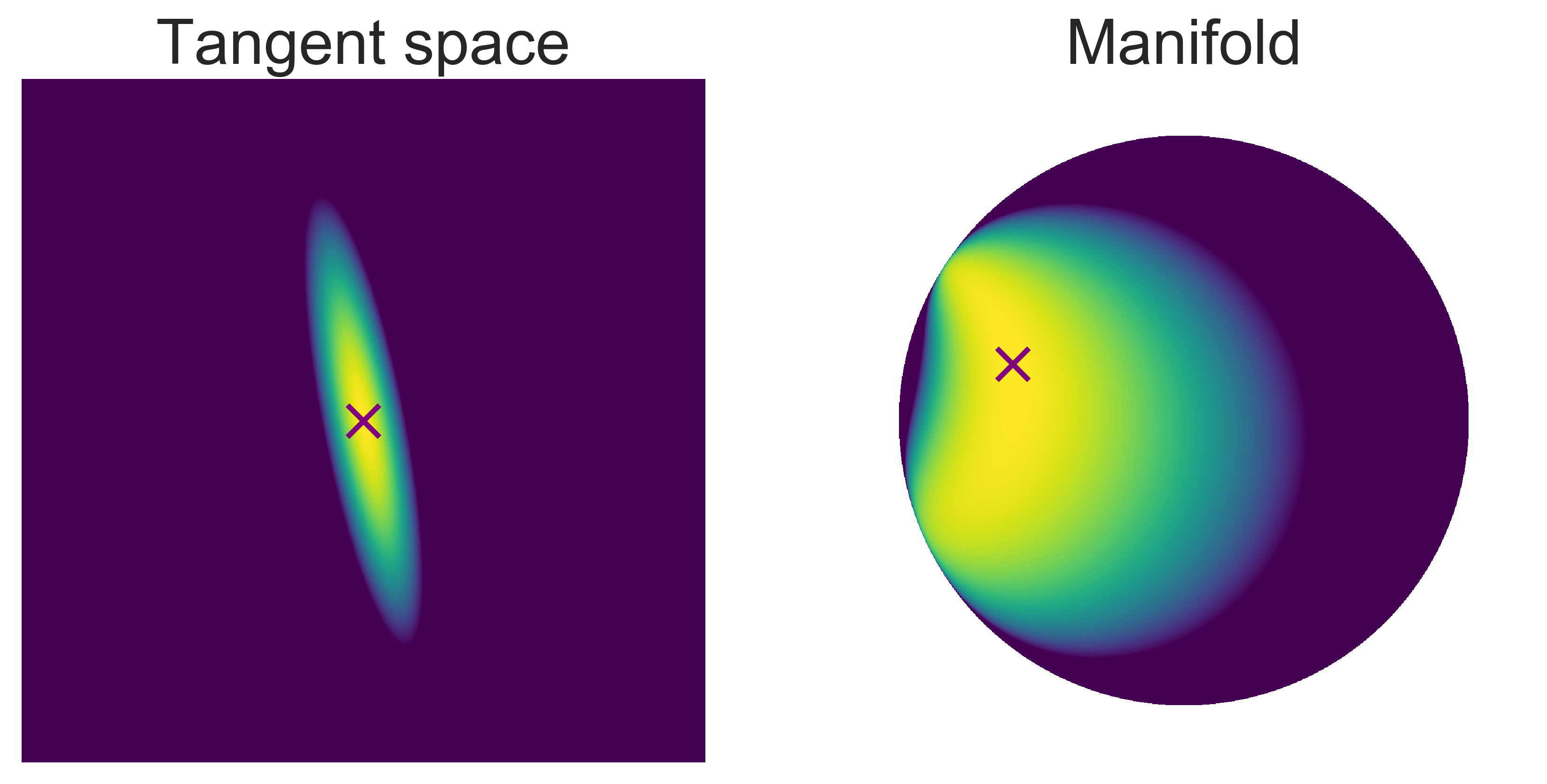}
    \end{minipage}

  \end{tabular}
  \caption{Visual examples of pseudo-hyperbolic Gaussian on $\mathbb{H}^2$.
  Log-density is illustrated on $\mathcal{B}^2$ by translating each point from $\mathbb{H}^2$ for clarity.
  We designate the origin of hyperbolic space by the $\times$ mark.}
  \label{fig:density_samples}
\end{figure}

Figure \ref{fig:density_samples} shows examples of pseudo-hyperbolic Gaussian $\mathcal{G}(\vb*{\mu}, \Sigma)$  with various $\mu$ and $\Sigma$.
We plotted the log-density of these distributions by heatmaps.
We designate the $\vb*{\mu}$ by the $\times$ mark.
The right side of these figures expresses their log-density on the Poincar\'e ball model, and the left side expresses the same one on the corresponding tangent space.

\vspace{-0.3cm}

\newpage

\section{Additional Numerical Evaluations}

\subsection{Synthetic Binary Tree}

\begin{figure*}[htbp]
  \centering
  \begin{tabular}{ccc}
    \begin{minipage}[t]{0.3\hsize}
      \centering
      \subcaption{A tree representation of the training dataset}
      \vspace{0.3cm}
      \includegraphics[width=\linewidth]{images/binary_tree/binary_tree_schema_v3.pdf}
      \label{fig:binary_tree_schema}
    \end{minipage}
    &
    \begin{minipage}[t]{0.3\hsize}
      \centering
      \subcaption{Vanilla ($\beta=0.1$)}
      \vspace{-0.1cm}
      \includegraphics[width=\linewidth]{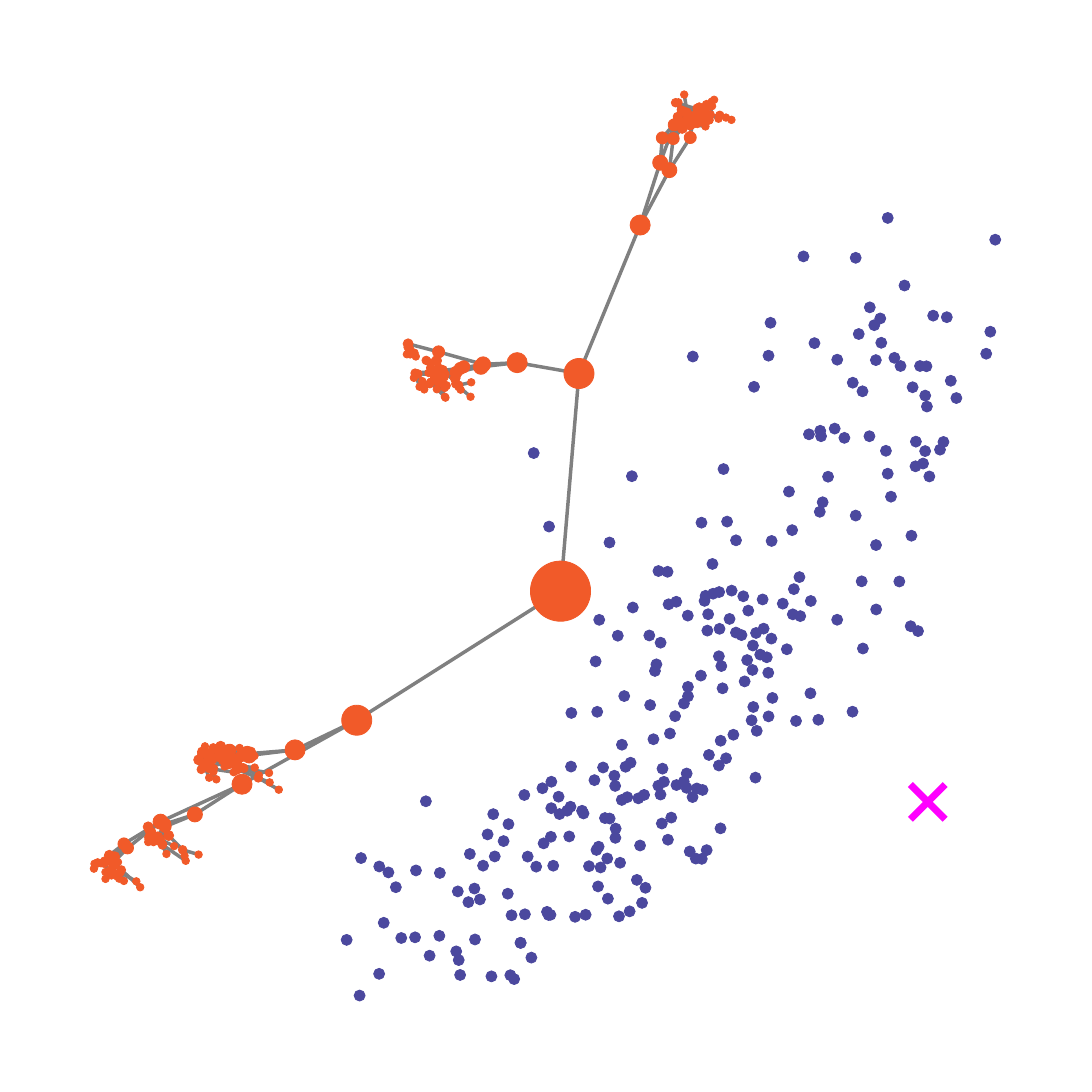}
      \label{fig:binary_tree_normal_beta1}
    \end{minipage}
    &
    \begin{minipage}[t]{0.3\hsize}
      \centering
      \subcaption{Vanilla ($\beta=1.0$)}
      \vspace{-0.1cm}
      \includegraphics[width=\linewidth]{images/binary_tree/embedding_normal.pdf}
      \label{fig:binary_tree_normal_beta1}
    \end{minipage}
    \\
    \begin{minipage}[t]{0.3\hsize}
      \centering
      \subcaption{Vanilla ($\beta=2.0$)}
      \vspace{-0.1cm}
      \includegraphics[width=\linewidth]{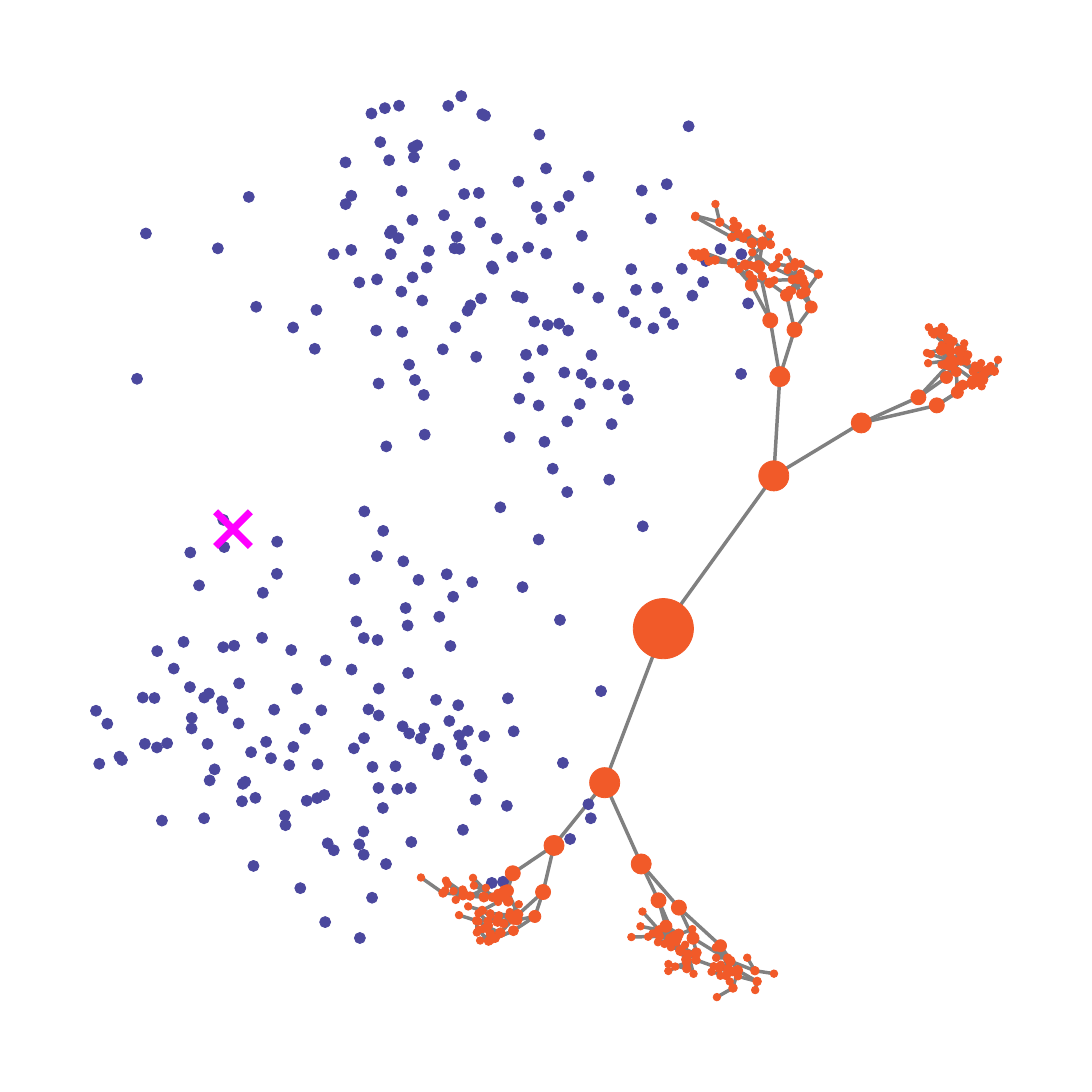}
      \label{fig:binary_tree_normal_beta2}
    \end{minipage}
    &
    \begin{minipage}[t]{0.3\hsize}
      \centering
      \subcaption{Vanilla ($\beta=3.0$)}
      \vspace{-0.1cm}
      \includegraphics[width=\linewidth]{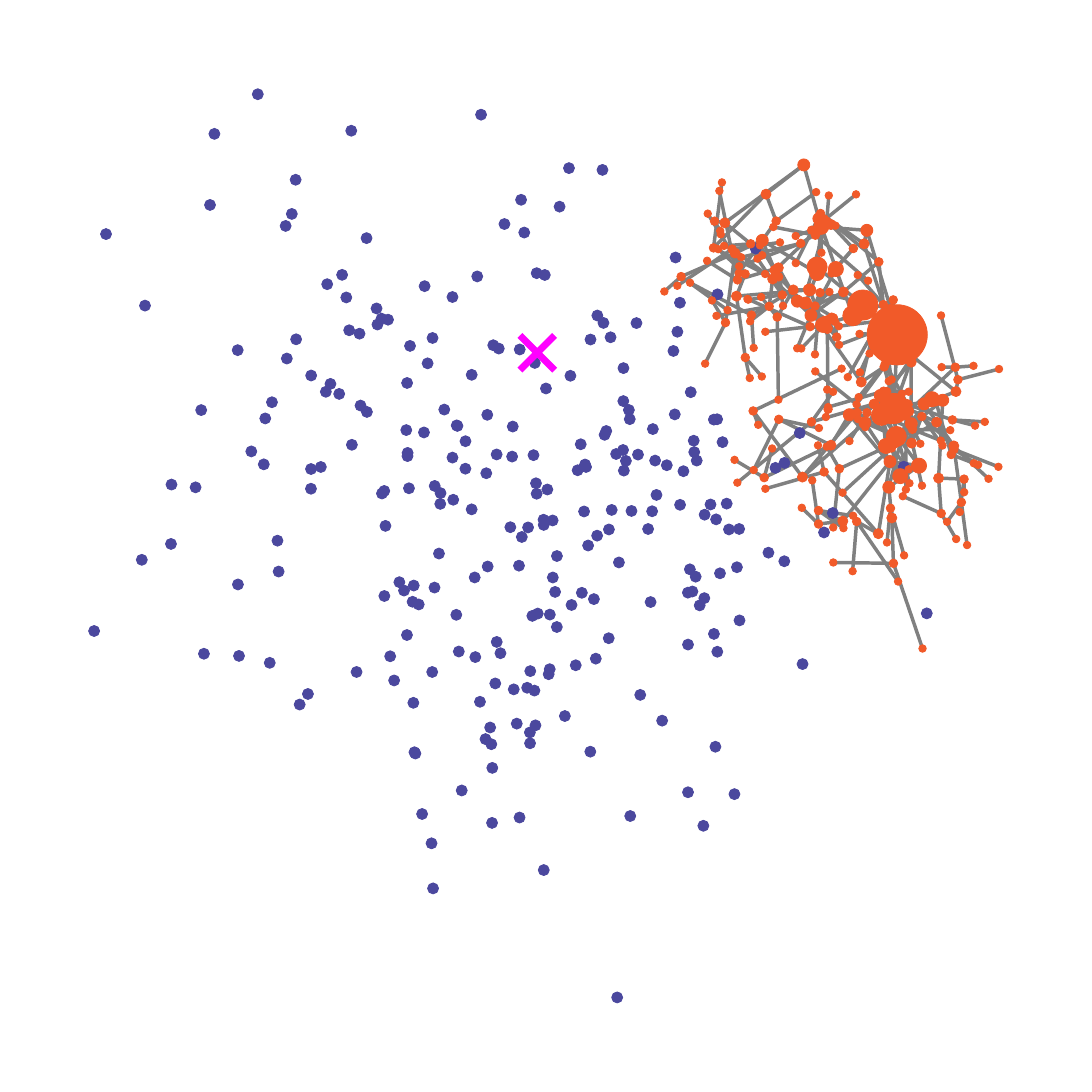}
      \label{fig:binary_tree_normal_beta3}
    \end{minipage}
    &
    \begin{minipage}[t]{0.3\hsize}
      \centering
      \subcaption{Hyperbolic}
      \vspace{-0.1cm}
      \includegraphics[width=\linewidth]{images/binary_tree/embedding_hyperbolic_wocircle.pdf}
      \label{fig:binary_tree_hyperbolic}
    \end{minipage}
  \end{tabular}
  \vspace{-0.7cm}
  \caption{The visual results of Vanilla and Hyperbolic VAEs applied to an artificial dataset generated by applying a random perturbation to a binary tree. The visualization is being done in the Poincar\'e ball. Red points are the embeddings of the original tree, and the blue points are the embeddings of all other points in the dataset. Pink $\times$ represents the origin of hyperbolic space. Note that the hierarchical relations in the original tree was \textbf{not} used during the training phase.}
  \label{fig:synthetic_tree_full_result}
\end{figure*}

We qualitatively compared the learned latent space of Vanilla and Hyperbolic VAEs.
Figure \ref{fig:synthetic_tree_full_result} shows the embedding vectors of the synthetic binary tree dataset on the two-dimensional latent space.
We evaluated the latent space of Vanilla VAE with $\beta = 0.1, 1.0, 2.0$, and $3.0$, and Hyperbolic VAE.
Note that the hierarchical relations in the original tree were \textbf{not} used during the training phase.
Red points are the embeddings of the noiseless observations.
As we mentioned in the main text, we evaluated the correlation coefficient between the Hamming distance on the data space and the hyperbolic (Euclidean for Vanilla VAEs) distance on the latent space.
Consistently with this metric, the latent space of the Hyperbolic VAE captured the hierarchical structure inherent in the dataset well.
In the comparison between Vanilla VAEs, the latent space captured the hierarchical structure according to increase the $\beta$.
However, the posterior distribution of the Vanilla VAE with $\beta = 3.0$ collapsed and lost the structure.
Also, the blue points are the embeddings of noisy observation, and pink $\times$ represents the origin of the latent space.
In latent space of Vanilla VAEs, there was bias in which embeddings of noisy observations were biased to the center side.

\newpage

\subsection{MNIST}

\begin{table}[htbp]
  \centering
  \begin{tabular}{lcccc}
    \toprule
    & \multicolumn{2}{c}{Vannila VAE} & \multicolumn{2}{c}{Hyperbolic VAE} \\
    \cmidrule(l){2-5}
    $n$ & ELBO & LL & ELBO & LL \\
    \midrule
    2  & $-145.53 {\scriptstyle \pm .65}$ & $-140.45 {\scriptstyle \pm .47}$ & $-143.23 {\scriptstyle \pm 0.63}$ & $\cellbest -138.61 {\scriptstyle \pm 0.45}$ \\
    5  & $-111.32 {\scriptstyle \pm .38}$ & $-105.78 {\scriptstyle \pm .51}$ & $-111.09 {\scriptstyle \pm 0.39}$ & $\cellbest -105.38 {\scriptstyle \pm 0.61}$ \\
    10  & $-92.49 {\scriptstyle \pm .52}$ & $\cellbest -86.25 {\scriptstyle \pm .52}$ & $-93.10 {\scriptstyle \pm 0.26}$ & $-86.40 {\scriptstyle \pm 0.28}$ \\
    20  & $-85.17 {\scriptstyle \pm .40}$ & $\cellbest -77.89 {\scriptstyle \pm .36}$ & $-88.28 {\scriptstyle \pm 0.34}$ & $-79.23 {\scriptstyle \pm 0.20}$ \\
    \bottomrule
  \end{tabular}
  \caption{Quantitative comparison of Hyperbolic VAE against Vanilla VAE on the MNIST dataset in terms of ELBO and log-likelihood (LL) for several values of latent space dimension $n$. LL was computed using 500 samples of latent variables. We calculated the mean and the $\pm 1$ SD with five different experiments.
  }
  \label{tab:mnist_result_full}
\end{table}

We showed the numerical performance of Vanilla and Hyperbolic VAEs for MNIST data in the main text in terms of the log-likelihood. In this section, we also show the evidence lower bound for the same dataset in Table \ref{tab:mnist_result_full}.

\subsection{Atari 2600 Breakout}

\begin{figure}[htbp]
  \centering
  \includegraphics[width=0.5\hsize]{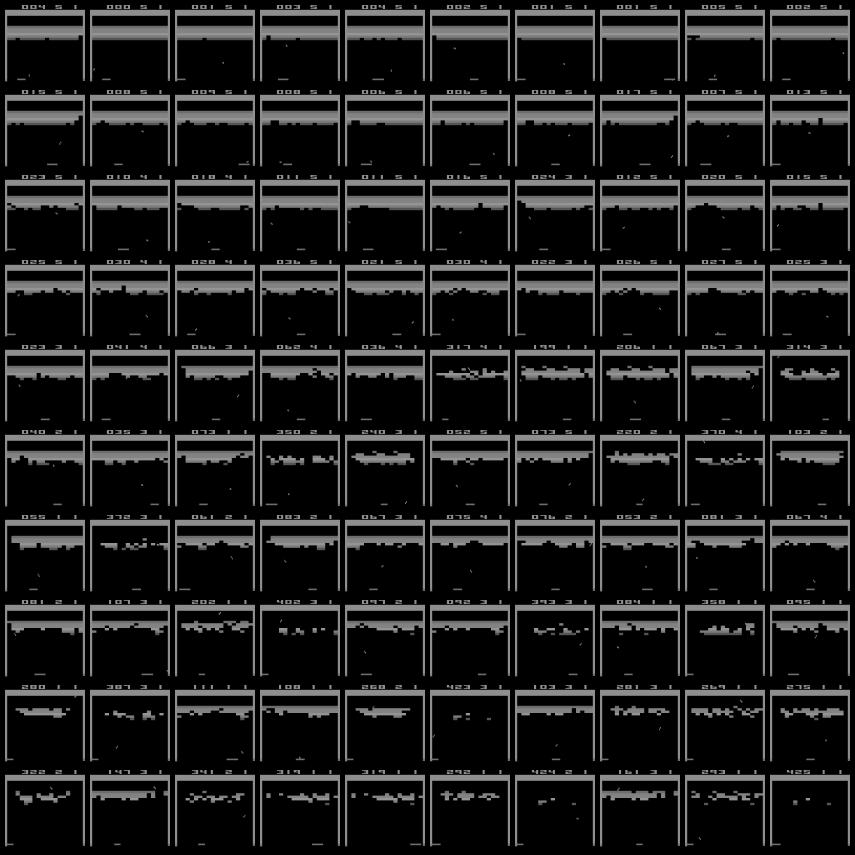}
  \caption{Examples of observed screens in Atari 2600 Breakout.}
  \label{fig:breakout_dataset_large}
\end{figure}

To evaluate the performance of Hyperbolic VAE for hierarchically organized dataset according to time development, we applied our Hyperbolic VAE to a set of trajectories that were explored by an agent with a trained policy during multiple episodes of Breakout in Atari 2600.
We used a pretrained Deep Q-Network to collect trajectories, and Figure \ref{fig:breakout_dataset_large} shows examples of observed screens.

\begin{figure*}[htbp]
  \centering
  \begin{tabular}{ccc}
    \begin{minipage}[t]{0.26\hsize}
      \centering
      \vspace{-0.5cm}
      \subcaption{$\norm{\tilde{\vb*{v}}}_2 = 0$}
      \includegraphics[width=\linewidth]{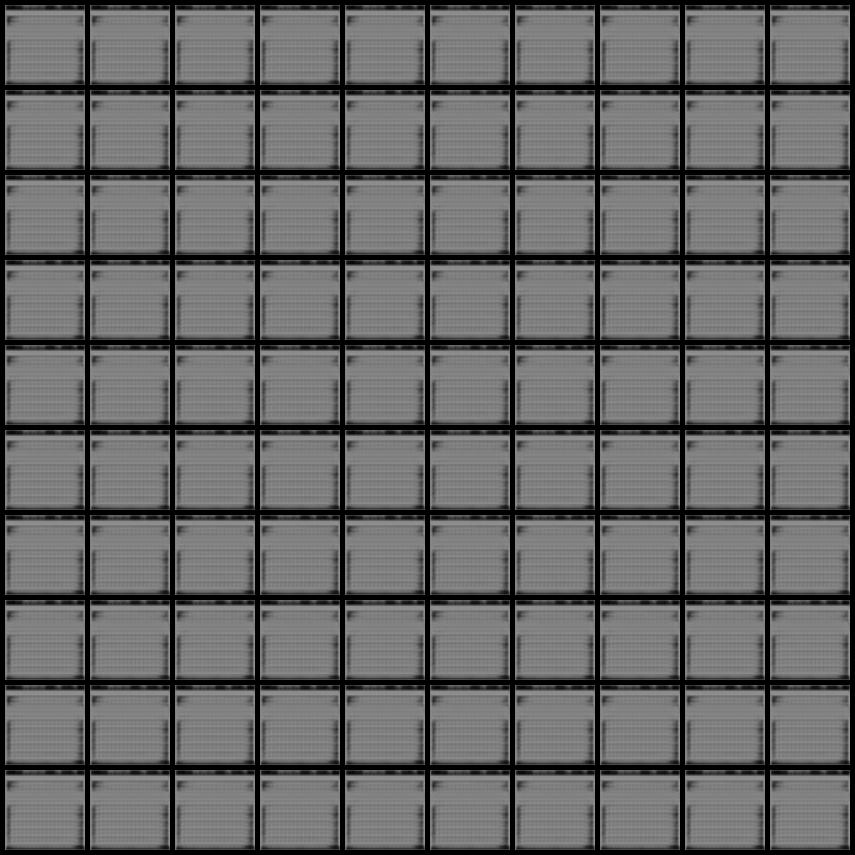}
      \label{fig:breakout_euclid_norm0}
    \end{minipage}
    &
    \begin{minipage}[t]{0.26\hsize}
      \centering
      \vspace{-0.5cm}
      \subcaption{$\norm{\tilde{\vb*{v}}}_2 = 1$}
      \includegraphics[width=\linewidth]{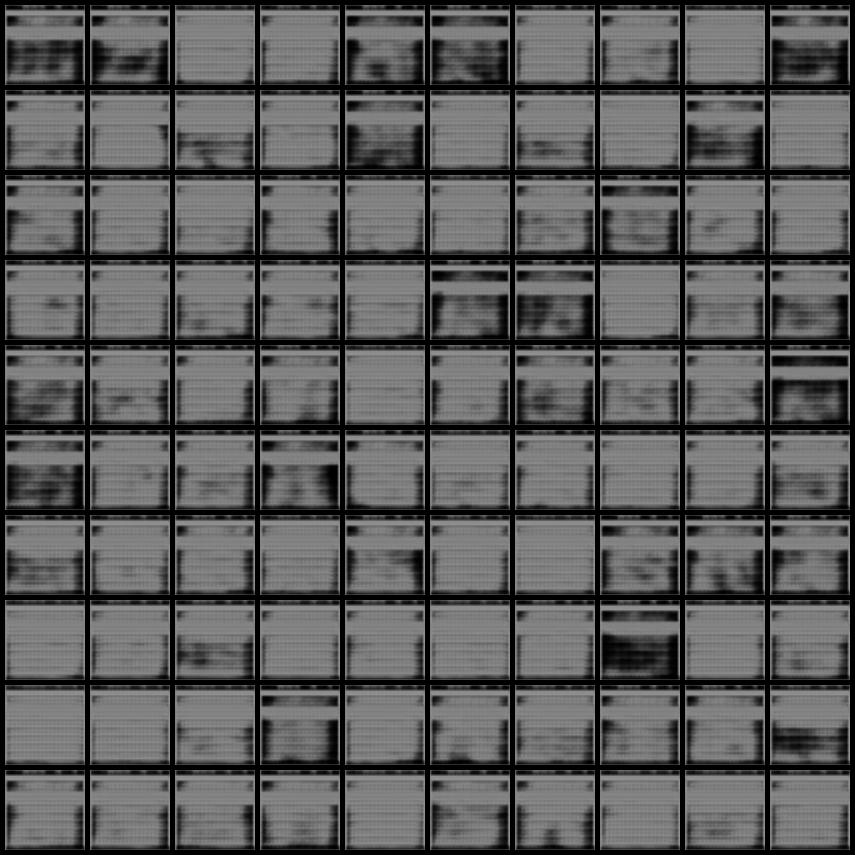}
      \label{fig:breakout_euclid_norm1}
    \end{minipage}
    &
    \begin{minipage}[t]{0.26\hsize}
      \centering
      \vspace{-0.5cm}
      \subcaption{$\norm{\tilde{\vb*{v}}}_2 = 2$}
      \includegraphics[width=\linewidth]{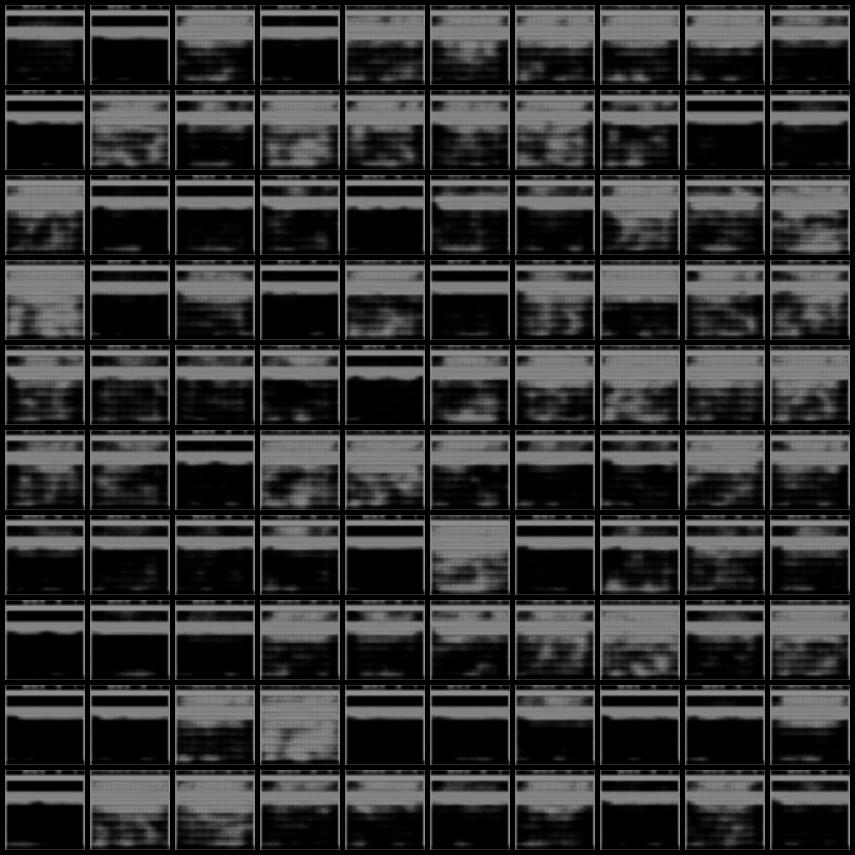}
      \label{fig:breakout_euclid_norm2}
    \end{minipage}
    \\
    \begin{minipage}[t]{0.26\hsize}
      \centering
      \vspace{-0.5cm}
      \subcaption{$\norm{\tilde{\vb*{v}}}_2 = 3$}
      \includegraphics[width=\linewidth]{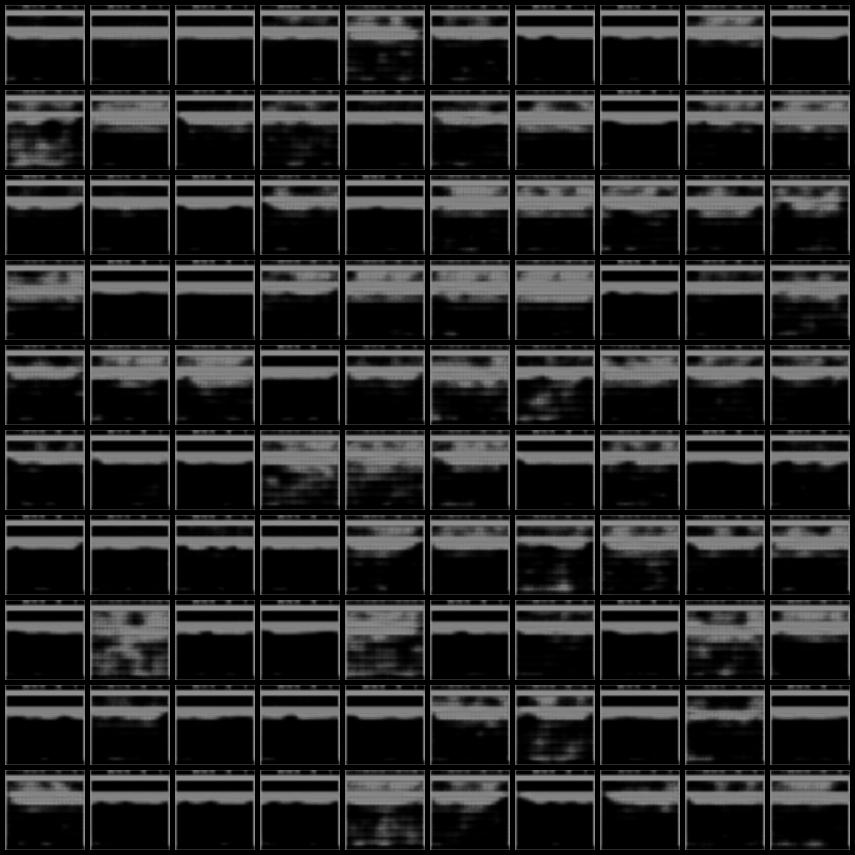}
      \label{fig:breakout_euclid_norm3}
    \end{minipage}
    &
    \begin{minipage}[t]{0.26\hsize}
      \centering
      \vspace{-0.5cm}
      \subcaption{$\norm{\tilde{\vb*{v}}}_2 = 5$}
      \includegraphics[width=\linewidth]{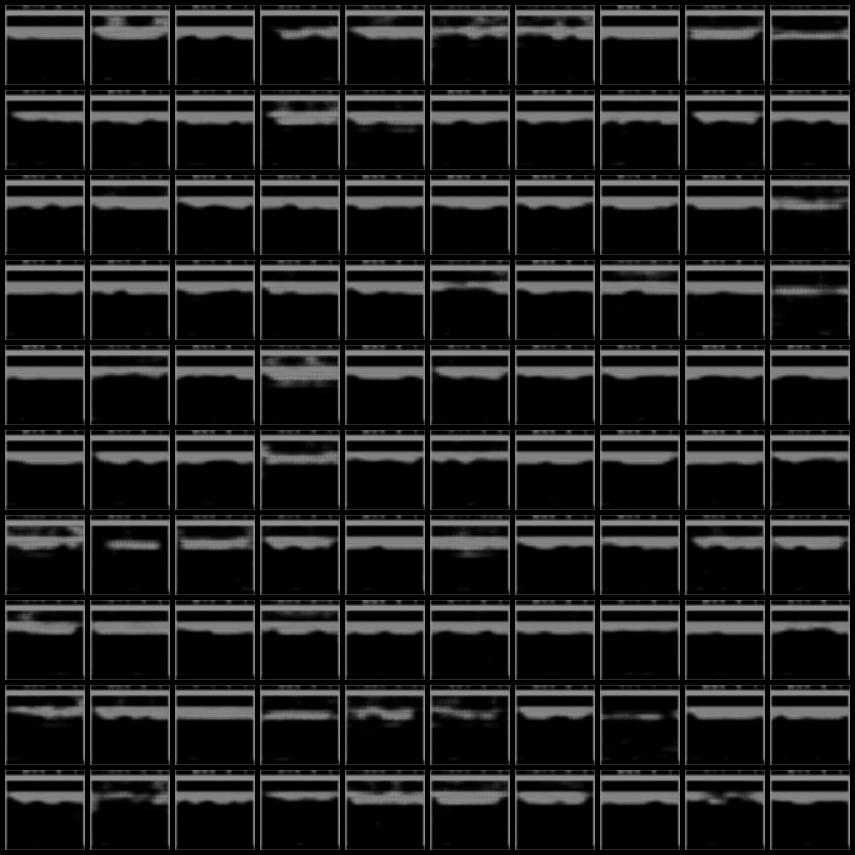}
      \label{fig:breakout_euclid_norm5}
    \end{minipage}
    &
    \begin{minipage}[t]{0.26\hsize}
      \centering
      \vspace{-0.5cm}
      \subcaption{$\norm{\tilde{\vb*{v}}}_2 = 10$}
      \includegraphics[width=\linewidth]{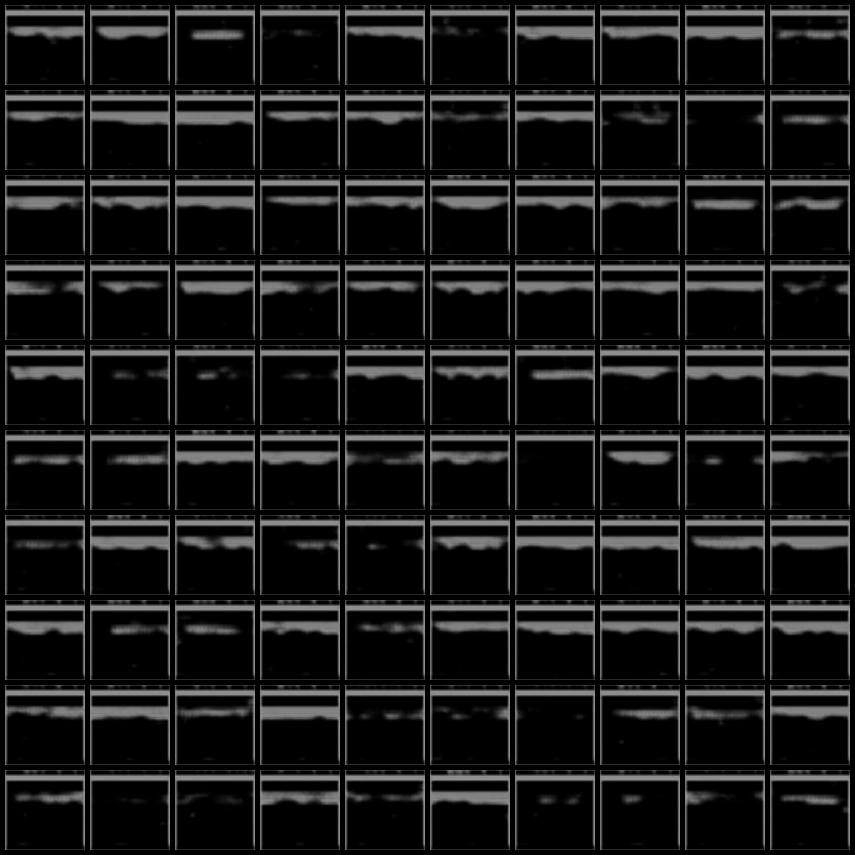}
      \label{fig:breakout_euclid_norm10}
    \end{minipage}
  \end{tabular}
  \vspace{-0.4cm}
  \caption{Images generated by Vanilla VAE with constant norm $\norm{\tilde{\vb*{v}}}_2=a$.}
  \label{fig:breakout_euclid_norm_sample}
\end{figure*}

\begin{figure*}[htbp]
  \centering
  \begin{tabular}{ccc}
    \begin{minipage}[t]{0.26\hsize}
      \centering
      \vspace{-0.5cm}
      \subcaption{$\norm{\tilde{\vb*{v}}}_2 = 0$}
      \includegraphics[width=\linewidth]{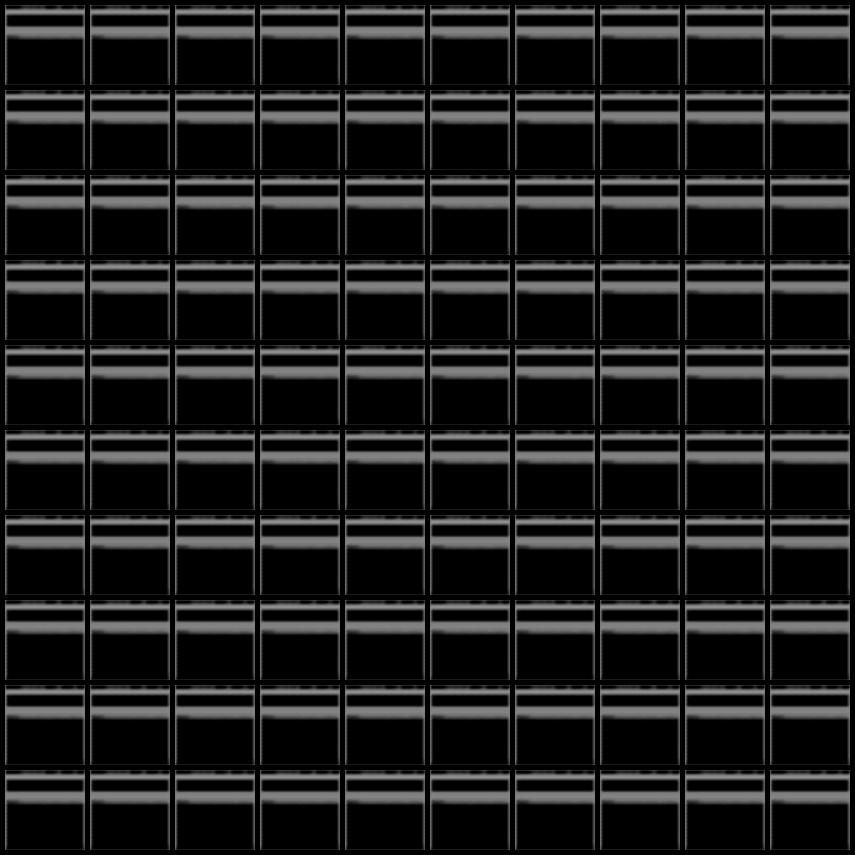}
      \label{fig:breakout_hyperbolic_norm0}
    \end{minipage}
    &
    \begin{minipage}[t]{0.26\hsize}
      \centering
      \vspace{-0.5cm}
      \subcaption{$\norm{\tilde{\vb*{v}}}_2 = 1$}
      \includegraphics[width=\linewidth]{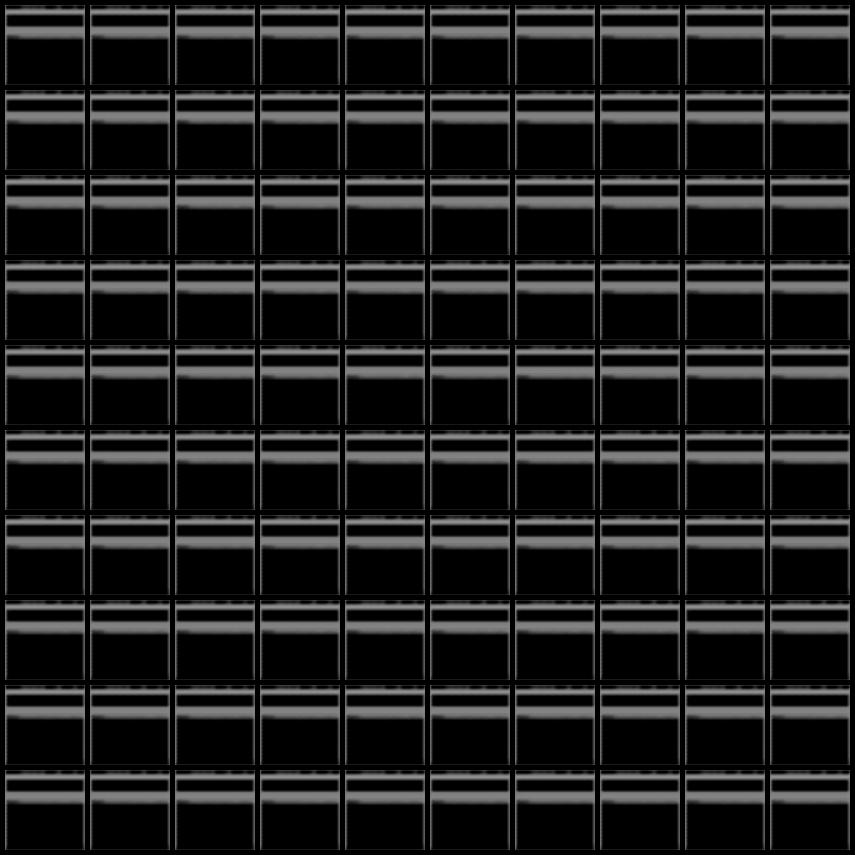}
      \label{fig:breakout_hyperbolic_norm1}
    \end{minipage}
    &
    \begin{minipage}[t]{0.26\hsize}
      \centering
      \vspace{-0.5cm}
      \subcaption{$\norm{\tilde{\vb*{v}}}_2 = 2$}
      \includegraphics[width=\linewidth]{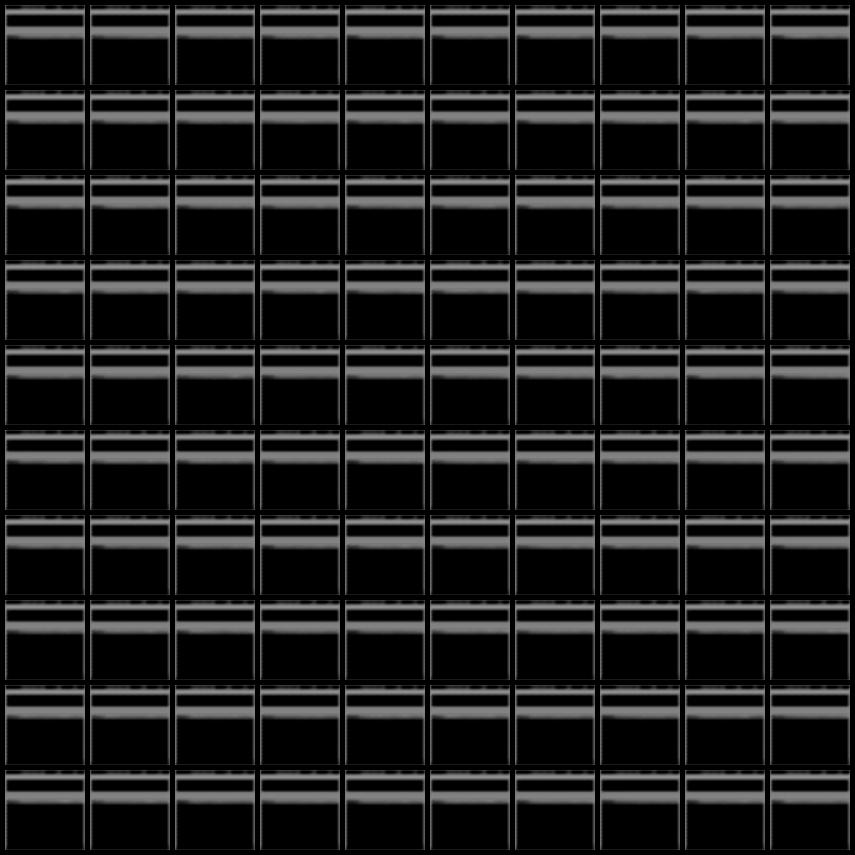}
      \label{fig:breakout_hyperbolic_norm2}
    \end{minipage}
    \\
    \begin{minipage}[t]{0.26\hsize}
      \centering
      \vspace{-0.5cm}
      \subcaption{$\norm{\tilde{\vb*{v}}}_2 = 3$}
      \includegraphics[width=\linewidth]{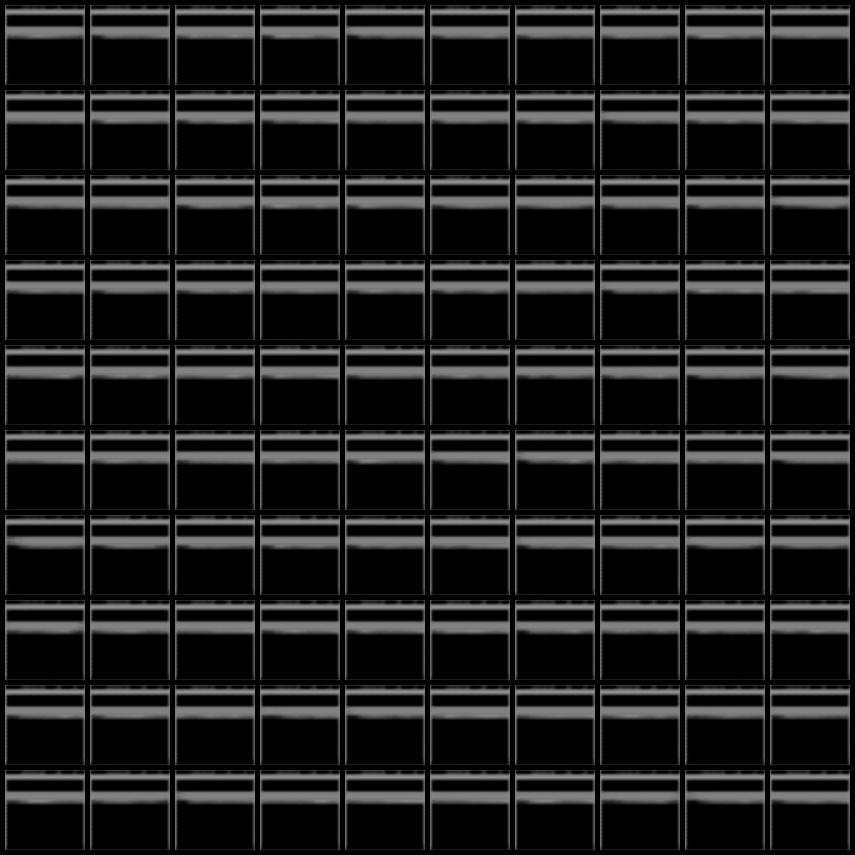}
      \label{fig:breakout_hyperbolic_norm3}
    \end{minipage}
    &
    \begin{minipage}[t]{0.26\hsize}
      \centering
      \vspace{-0.5cm}
      \subcaption{$\norm{\tilde{\vb*{v}}}_2 = 5$}
      \includegraphics[width=\linewidth]{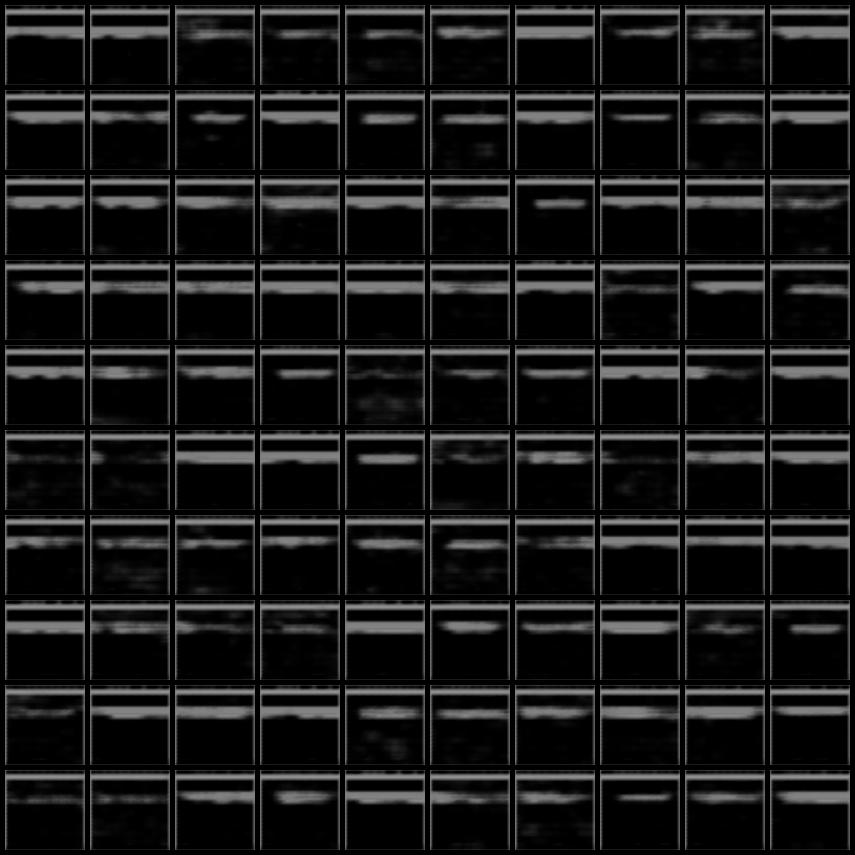}
      \label{fig:breakout_hyperbolic_norm5}
    \end{minipage}
    &
    \begin{minipage}[t]{0.26\hsize}
      \centering
      \vspace{-0.5cm}
      \subcaption{$\norm{\tilde{\vb*{v}}}_2 = 10$}
      \includegraphics[width=\linewidth]{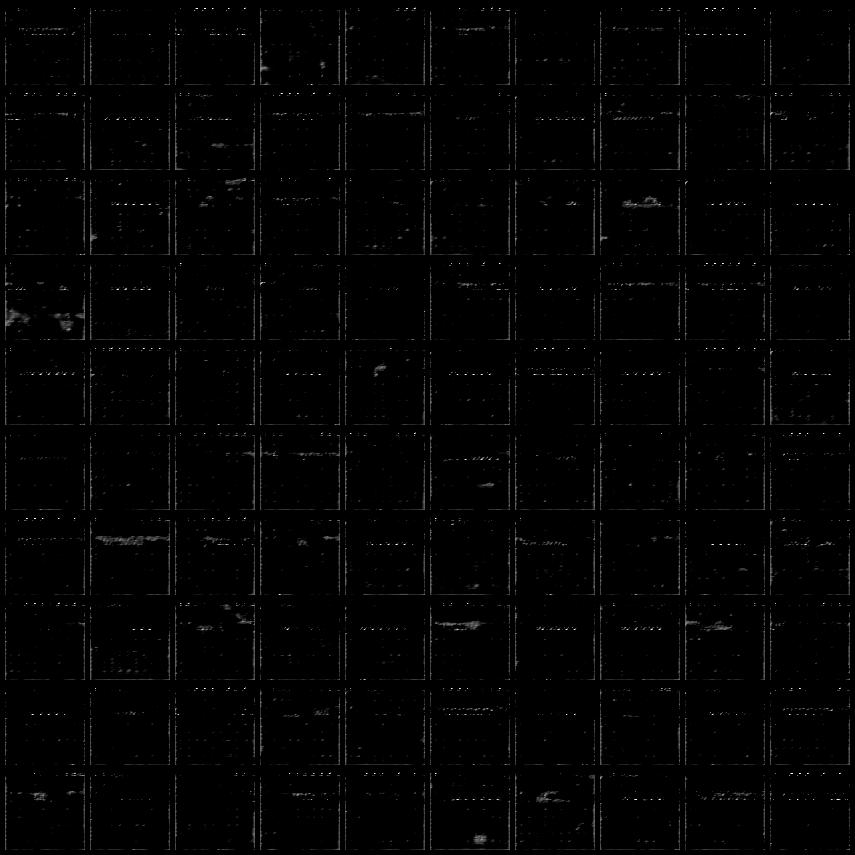}
      \label{fig:breakout_hyperbolic_norm10}
    \end{minipage}
  \end{tabular}
  \vspace{-0.4cm}
  \caption{Images generated by Hyperbolic VAE with constant norm $\norm{\tilde{\vb*{v}}}_2=a$.}
  \label{fig:breakout_hyperbolic_norm_sample}
\end{figure*}

We showed three trajectories of samples from the prior distribution with the scaled norm for both models in the main text.
We also visualize more samples in Figure \ref{fig:breakout_euclid_norm_sample} and \ref{fig:breakout_hyperbolic_norm_sample}.
For both models, we generated samples with $\norm{\tilde{\vb*{v}}}_2 =$ 0, 1, 2, 3, 5, and 10.

Vanilla VAE tended to generate oversaturated images when the norm $\norm{\tilde{\vb*{v}}}$ was small.
Although the model generated several images which include a small number of blocks as the norm increases, it also generated images with a constant amount of blocks even $\norm{\tilde{\vb*{v}}} = 10$.
On the other hand, the number of blocks contained in the generated image of Hyperbolic VAE gradually decreased according to the norm.

\newpage

\subsection{Word Embeddings}

We showed the experimental results of probabilistic word embedding models with diagonal variance in the main text.
Table \ref{tab:wordnet_result_full} shows the same comparison with the reference model by \citet{Nickel2017}.
We also show the results with unit variance (Table \ref{tab:wordnet_result_circular}).
When the dimensions of the latent variable are small, the performance of the model on hyperbolic space did not deteriorate much by changing the variance from diagonal to unit.
However, the same change dramatically worsened the performance of the model on Euclidean space.

\begin{table}[htbp]
  \centering
  \begin{tabular}{lcccc|cc}
    \toprule
    & \multicolumn{2}{c}{Euclid} & \multicolumn{2}{c}{Hyperbolic} &
    \multicolumn{2}{c}{\citet{Nickel2017}} \\
    \cmidrule(l){2-7}
    $n$ & MAP & Rank & MAP & Rank & MAP & Rank \\
    \midrule
    5   & $0.296 {\scriptstyle \pm .006}$ & $25.09 {\scriptstyle \pm .80}$ & $\cellbest 0.506 {\scriptstyle \pm .017}$ & $\cellbest 20.55 {\scriptstyle \pm 1.34}$ & 0.823 & 4.9 \\
    10  & $0.778 {\scriptstyle \pm .007}$ & $\cellbest 4.70 {\scriptstyle \pm .05}$ & $\cellbest 0.795 {\scriptstyle \pm .007}$ & $5.07 {\scriptstyle \pm .12}$ & 0.851 & 4.02  \\
    20  & $0.894 {\scriptstyle \pm .002}$ & $\cellbest 2.23 {\scriptstyle \pm .03}$ & $\cellbest 0.897 {\scriptstyle \pm .005}$ & $2.54 {\scriptstyle \pm .20}$ & 0.855 & 3.84  \\
    50  & $0.942 {\scriptstyle \pm .003}$ & $1.51 {\scriptstyle \pm .04}$ & $\cellbest 0.975 {\scriptstyle \pm .001}$ & $\cellbest 1.19 {\scriptstyle \pm .01}$ & 0.86 & 3.98  \\
    100 & $0.953 {\scriptstyle \pm .002}$ & $1.34 {\scriptstyle \pm .02}$ & $\cellbest 0.978 {\scriptstyle \pm .002}$ & $\cellbest 1.15 {\scriptstyle \pm .01}$ & 0.857 & 3.9 \\
    \bottomrule
  \end{tabular}
  \caption{Experimental results of the reconstruction performance on the transitive closure of the WordNet noun hierarchy for several latent space dimension $n$. We calculated the mean and the $\pm 1$ SD with three different experiments.
  }\label{tab:wordnet_result_full}
\end{table}

\begin{table}[htbp]
  \centering
  \begin{tabular}{lcccc}
    \toprule
    & \multicolumn{2}{c}{Euclid} & \multicolumn{2}{c}{Hyperbolic} \\
    \cmidrule(l){2-5}
    $n$ & MAP & Rank & MAP & Rank \\
    \midrule
    5   & $0.217 {\scriptstyle \pm .008}$ & $55.28 {\scriptstyle \pm 3.54}$ & $\cellbest 0.529 {\scriptstyle \pm .010}$ & $\cellbest 22.38 {\scriptstyle \pm .70}$ \\
    10  & $0.698 {\scriptstyle \pm .030}$ & $6.54 {\scriptstyle \pm .65}$ & $\cellbest 0.771 {\scriptstyle \pm .006}$ & $\cellbest 5.89 {\scriptstyle \pm .29}$  \\
    20  & $0.832 {\scriptstyle \pm .016}$ & $3.08 {\scriptstyle \pm .16}$ & $\cellbest 0.862 {\scriptstyle \pm .002}$ & $\cellbest 2.80 {\scriptstyle \pm .13}$  \\
    50  & $\cellbest 0.910 {\scriptstyle \pm .006}$ & $\cellbest 1.78 {\scriptstyle \pm .071}$ & $0.903 {\scriptstyle \pm .003}$ & $1.94 {\scriptstyle \pm .03}$  \\
    100 & $0.882 {\scriptstyle \pm .008}$ & $4.75 {\scriptstyle \pm 2.01}$ & $\cellbest 0.884 {\scriptstyle \pm .003}$ & $\cellbest 2.57 {\scriptstyle \pm .09}$ \\
    \bottomrule
  \end{tabular}
  \caption{Experimental results of the word embedding models with unit variance on the WordNet noun dataset. We calculated the mean and the $\pm 1$ SD with three different experiments.
  }
  \label{tab:wordnet_result_circular}
\end{table}

\newpage

\section{Network Architecture}

Table \ref{tab:network_architecture} shows the network architecture that we used in Breakout experiments.
We evaluated Vanilla and Hyperbolic VAEs with a DCGAN-based architecture \cite{Radford2015} with the kernel size of the convolution and deconvolution layers as 3.
We used leaky ReLU nonlinearities for the encoder and ReLU nonlinearities for the decoder.
We set the latent space dimension as 20.
We gradually increased $\beta$ from 0.1 to 4.0 linearly during the first 30 epochs.
To ensure the initial embedding vector close to the origin, we initialized $\gamma$ for the batch normalization layer \cite{Ioffe2015} of the encoder as 0.1.
We modeled the probability distribution of the data space $p(\vb*{x} | \vb*{z})$ as Gaussian, so the decoder output a vector twice as large as the original image.

\begin{table}[htbp]
  \centering
  \small
  \begin{tabular}{cc}
  \begin{minipage}[t]{0.45\hsize}
    \centering
    \begin{tabular}{ll}
      \toprule
      \multicolumn{2}{l}{Encoder} \\
      \cmidrule(l){1-2}
      Layer & Size \\
      \midrule
      Input & $80 \times 80 \times 1$ \\
      Convolution & $80 \times 80 \times 16$ \\
      BatchNormalization & \\
      Convolution & $40 \times 40 \times 32$ \\
      BatchNormalization & \\
      Convolution & $40 \times 40 \times 32$ \\
      BatchNormalization & \\
      Convolution & $20 \times 20 \times 64$ \\
      BatchNormalization & \\
      Convolution & $20 \times 20 \times 64$ \\
      BatchNormalization & \\
      Convolution & $10 \times 10 \times 64$ \\
      Linear & $2n$ \\
      \bottomrule
    \end{tabular}
  \end{minipage}
  &
  \begin{minipage}[t]{0.45\hsize}
    \centering
    \begin{tabular}{ll}
      \toprule
      \multicolumn{2}{l}{Decoder} \\
      \cmidrule(l){1-2}
      Layer & Size \\
      \midrule
      Linear & $10 \times 10 \times 64$ \\
      BatchNormalization & \\
      Deconvolution & $20 \times 20 \times 32$ \\
      BatchNormalization & \\
      Convolution & $20 \times 20 \times 32$ \\
      BatchNormalization & \\
      Deconvolution & $40 \times 40 \times 16$ \\
      BatchNormalization & \\
      Convolution & $40 \times 40 \times 16$ \\
      Deconvolution & $80 \times 80 \times 2$ \\
      Convolution & $80 \times 80 \times 2$ \\
      \bottomrule
    \end{tabular}
  \end{minipage}
  \end{tabular}
  \caption{Network architecture for Atari 2600 Breakout dataset.}
  \label{tab:network_architecture}
\end{table}

\end{document}